\documentclass{article} 
\usepackage{iclr2027_conference,times}

\usepackage{amsmath,amsfonts,bm}

\def\eqref#1{equation~\ref{#1}}

\def\1{\bm{1}}

\DeclareMathAlphabet{\mathsfit}{\encodingdefault}{\sfdefault}{m}{sl}
\SetMathAlphabet{\mathsfit}{bold}{\encodingdefault}{\sfdefault}{bx}{n}

\usepackage[hidelinks]{hyperref}
\usepackage{url}
\usepackage{amsmath}
\usepackage{amssymb}
\usepackage{graphicx}
\usepackage{wrapfig}
\usepackage{multirow}

\usepackage{algorithm}
\usepackage{algorithmic}
\usepackage[inline]{enumitem}
\usepackage{enumitem}
\usepackage{boldline}
\usepackage{hhline}
\usepackage{subcaption}
\usepackage{array}
\usepackage{booktabs}
\usepackage[table]{xcolor}

\title{Stability–Plasticity Balance via Singular-Vector Selection in LLM Continual Learning}

\author{
Lingxiang Wang\textsuperscript{1,3}\thanks{Email: \texttt{wanglingxiang@buaa.edu.cn}},
Hainan Zhang\textsuperscript{1,3}\thanks{Corresponding author. Email: \texttt{zhanghainan@buaa.edu.cn}},
Liang Pang\textsuperscript{2},
Hongwei Zheng\textsuperscript{3},
Zhiming Zheng\textsuperscript{1,3} \\
\textsuperscript{1}School of Artificial Intelligence, Beihang University \\
\textsuperscript{2}Institute of Computing Technology, Chinese Academy of Sciences \\
\textsuperscript{3}Beijing Advanced Innovation Center for Future Blockchain and Privacy Computing
}

\iclrfinalcopy 
\begin{document}

\maketitle

\begin{abstract}
Domain-specific continual adaptation of LLMs risks catastrophic forgetting, creating a fundamental tension between acquiring new capabilities and preserving those learned during pretraining. 
PEFT mitigates this problem by restricting the number of trainable parameters,
but existing methods lack a principled unit for deciding where plasticity should be allocated and stability should be preserved. We identify the singular-vector channel as a natural unit for managing this trade-off. Each channel represents an input–output transformation, which can be updated to acquire new knowledge or fixed to preserve pretrained capabilities. Based on this perspective, we introduce SVC, a parameter-efficient continual-learning method that selectively updates Singular-Vector Channels. Before fine-tuning, 
SVC uses domain-specific data to estimate each channel’s adaptation benefit and a fixed public general-domain corpus only as a history activation proxy for estimating forgetting cost.
It then adaptively selects trainable channels based on these scores via knee-based cost screening, Pareto-front filtering, and Otsu thresholding.
Experimental results across four LLM families and eight downstream tasks show that SVC better preserves pretrained capabilities while achieving strong downstream performance relative to existing PEFT baselines. Further analysis of channel scoring and selection demonstrates that selective plasticity at the singular-vector-channel level enables effective continual LLM adaptation.

\end{abstract}

\section{Introduction}
\begin{wrapfigure}{r}{0.44\textwidth}
    \vspace{-15pt}
    \includegraphics[
        width=\linewidth
    ]{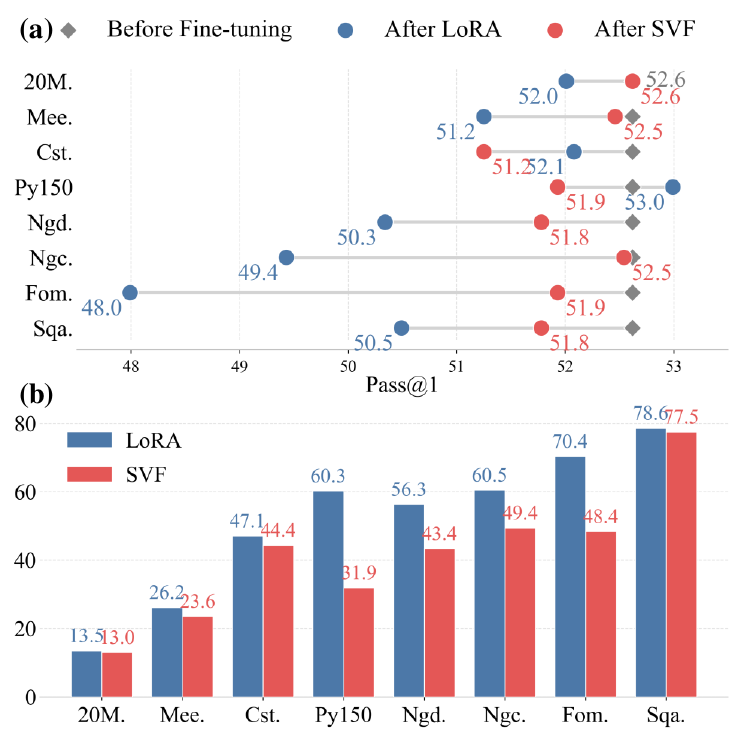}
    \vspace{-20pt}
    \caption{\textbf{Stability--plasticity trade-off between LoRA and SVF.} (a) GSM8K retention across fine-tuning tasks. (b) Downstream performance. Details in Appendix~C}
    \label{fig:motivation}
    \vspace{-15pt}
\end{wrapfigure}
Large language models (LLMs) gain broad knowledge and general-purpose reasoning through large-scale pretraining~\citep{karanikolas2023large}, but often require domain-specific fine-tuning to meet specialized data and task demands~\citep{chen2025mimo}. However, such adaptation can cause catastrophic forgetting, degrading pretrained capabilities and reducing reliability in real-world settings~\citep{french1999catastrophic,liu2020learning}.
Therefore, continual domain adaptation requires a careful balance between \emph{plasticity}, to acquire new domain knowledge, and \emph{stability}, to preserve pretrained capabilities. 

Traditional continual learning methods~\citep{shi2024continual} mitigate forgetting primarily by preserving information from previous learning stages. Regularization-based approaches constrain parameter updates or functional changes to protect previously acquired knowledge~\citep{kirkpatrick2017overcoming,li2017learning}, while replay-based approaches revisit stored or generated samples from earlier tasks~\citep{rebuffi2017icarl,rolnick2019experience}. Applying these strategies to LLMs is challenging because the original pretraining data and training signals are typically unavailable, making it difficult to directly preserve the broad capabilities acquired during pretraining. Moreover, overly strong constraints can impede adaptation to new domains, whereas weak constraints may fail to prevent forgetting. As a result, achieving a favorable stability-plasticity trade-off remains challenging for continual domain adaptation of LLMs.

Parameter-efficient fine-tuning (PEFT) offers a promising alternative by freezing most pretrained parameters and adapting only a small subset~\citep{ding2023parameter,10889361}. A prominent example is low-rank adaptation (LoRA)~\citep{hu2022lora}, which learns low-rank weight updates to enable effective downstream adaptation while often reducing degradation of pretrained capabilities~\citep{biderman2024lora}. However, the extent of such preservation can vary substantially across adaptation tasks. As shown in Figure~\ref{fig:motivation}(a), when the same pretrained model is independently adapted to each of the eight TRACE tasks~\citep{wang2023trace} using LoRA, the resulting models exhibit markedly different drops in GSM8K performance~\citep{cobbe2021training}. This suggests that restricting the update rank alone does not provide consistent control over forgetting.

Singular value fine-tuning (SVF)~\citep{sun2022singular} imposes a stronger structural constraint. Given a weight matrix decomposed by singular value decomposition, SVF updates only the singular values while keeping the corresponding left and right singular vectors fixed. Since the singular vectors determine the input and output directions of each rank-one component, whereas the singular values control their strengths~\citep{muller2004singular}, SVF modifies the magnitude of existing transformations without changing their directions. As shown in Figure~\ref{fig:motivation}(a), this stronger directional constraint yields substantially more consistent retention of GSM8K performance across tasks. However, Figure~\ref{fig:motivation}(b) shows that it also limits downstream adaptation performance.

These observations reveal a central tension in parameter-efficient adaptation: allowing transformation directions to change improves plasticity but can increase interference with pretrained capabilities, while fixing all directions improves stability at the cost of adaptation. This suggests that the key question is not whether transformation directions should be updated, but \emph{which directions should be updated for a given task}. To enable such task-dependent control, we treat each paired left and right singular-vector direction associated with a singular value as a \emph{singular-vector channel}, which serves as the basic unit for balancing adaptation benefit against forgetting cost.

In this paper, we introduce \textbf{SVC}, a parameter-efficient continual adaptation method that selectively updates Singular-Vector Channels. Before fine-tuning, SVC evaluates each channel using two task-dependent signals: its potential benefit for learning the current domain and its potential cost to previously acquired or general-domain capabilities. 
SVC then identifies candidate channels through knee-based cost screening and Pareto-front filtering in the benefit–cost plane, and applies Otsu thresholding to automatically select a task-specific subset for optimization.
During adaptation, all singular values and the singular vectors of the selected channels are updated jointly, while the remaining singular-vector directions are kept fixed. In this way, SVC retains the structural stability of singular value fine-tuning while introducing directional flexibility only where it is beneficial for the current task.


Extensive experiments across four LLM families and eight downstream tasks show that SVC effectively adapts to sequential tasks while consistently preserving pretrained general capabilities. Further analyses validate the effectiveness of channel-level benefit-cost estimation and the multi-stage selection mechanism, showing that selective channel updates better balance model stability and plasticity. 
Our contributions are summarized as follows:
\begin{itemize}
\item We identify singular-vector channels as a principled unit for controlling the stability-plasticity trade-off in parameter-efficient continual adaptation.

\item We propose a parameter-efficient continual-learning method for LLMs that adaptively balances stability and plasticity across tasks.

\item SVC requires neither prior-task samples nor task-specific modules, yet consistently performs well across models and task orders, demonstrating strong generality and scalability.

\end{itemize}
\section{Related Work}
{Parameter-Efficient Fine-Tuning} (PEFT) adapts LLMs through a small number of trainable parameters while keeping most pretrained weights fixed~\citep{ding2023parameter}.
Recent studies have shown that this restricted update regime can reduce catastrophic forgetting~\citep{vu2022overcoming,wang2025parameter}, with LoRA exhibiting particularly favorable retention behavior~\citep{biderman2024lora}.
O-LoRA~\citep{wang2023orthogonal}, a representative LoRA-based continual-learning method, learns successive tasks in orthogonal low-rank subspaces to reduce interference.
However, its orthogonality constraint is defined only relative to the low-rank subspaces learned for previous tasks.
It does not directly estimate whether changing a particular direction benefits the current domain or compromises pretrained general capabilities.
Spectral PEFT methods provide another way to constrain adaptation by exploiting the singular-value structure of pretrained weights.
SVF~\citep{sun2022singular}, SVDiff~\citep{han2023svdiff}, and SAM-PARSER~\citep{peng2024sam} keep the left and right singular-vector bases fixed and optimize only the diagonal SVD coefficients, thereby rescaling the existing matched rank-one components.
SVFT~\citep{lingam2024svft} instead learns a sparse coefficient matrix over fixed singular-vector bases, including off-diagonal coefficients that combine different left and right singular vectors.
Across these methods, the set of trainable spectral coefficients is specified before fine-tuning rather than selected using task-specific estimates of adaptation benefit and forgetting cost.
SVC instead treats singular-vector channels as the basic units for weighing domain-adaptation benefits against general-capability forgetting costs and adaptively selects trainable directions for each domain.
\section{Method}
\begin{figure}[t]
\vspace{-10pt}
    \begin{center}
    \includegraphics[width=\textwidth]{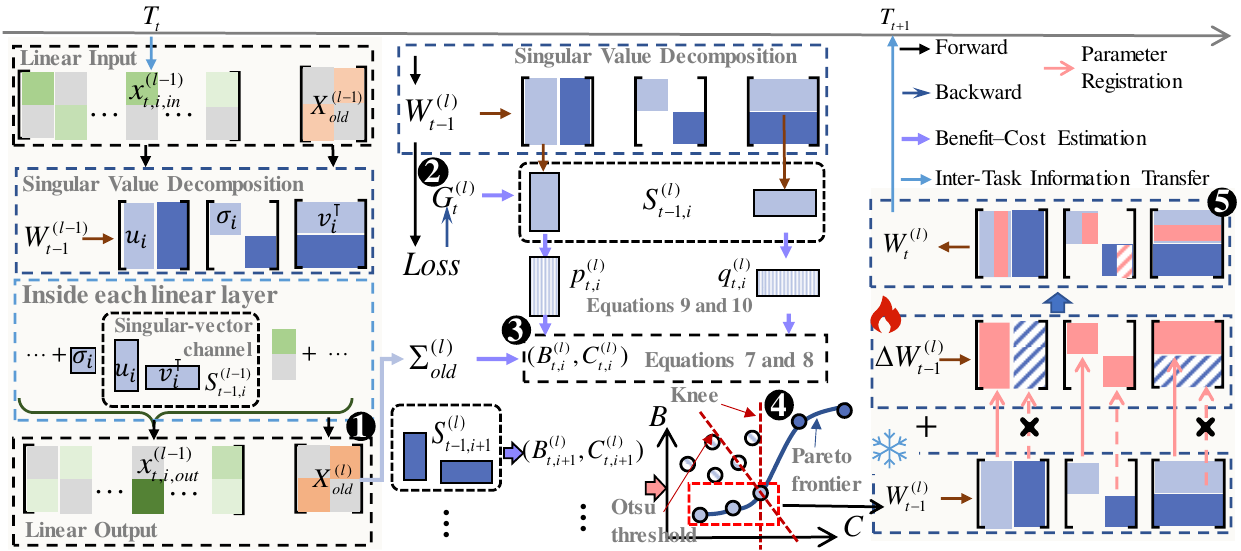}
    \end{center}
    \vspace{-5pt}
    \caption{\textbf{Overview of SVC.}
    SVC \textcircled{1} collects reference activation statistics from a publicly available, unlabeled general-domain reference corpus.
    It then \textcircled{2} computes layer-wise average gradients from current-task data.
    From these quantities, SVC \textcircled{3} estimates each singular-vector channel's local adaptation benefit and forgetting cost.
    SVC next \textcircled{4} selects trainable channels through knee-based cost screening, Pareto-front filtering, and Otsu thresholding.
    Finally, SVC \textcircled{5} updates all singular values while updating singular vectors only in selected channels.
    Post-training, SVC merges the learned increments into the model weights.}
    \vspace{-10pt}
    \label{fig:pipeline}
\end{figure}
This section presents SVC, covering the preliminaries, the estimation of adaptation benefit and forgetting cost for singular-vector channels, Pareto-frontier-based adaptive channel selection, and the overall fine-tuning procedure.
As illustrated in Figure~\ref{fig:pipeline}, singular value decomposition represents each linear-layer weight matrix as a collection of rank-one components, each defined by a singular value and a corresponding pair of left and right singular vectors.

\subsection{Preliminaries}
Continual learning requires a model to learn a sequence of tasks while preserving previously acquired capabilities.
Consider a task sequence \(\mathcal{T}=\{T_1,T_2,\ldots,T_T\}\), where task \(T_t\) is associated with a dataset \(\mathcal{D}_t=\{({x}_{t,i},{y}_{t,i})\}_{i=1}^{n_t}\).
Let \({\theta}\) denote the parameters of the model \(f(\cdot;{\theta})\). When training on task \(T_t\), the model is initialized with \({\theta}_{t-1}\), obtained after the preceding task, and updates its parameters by minimizing the current-task loss:
\begin{equation}\theta_t=\arg\min_{\theta}\frac{1}{n_t}\sum_{j=1}^{n_t}L_t(f({x}_{t,j};\theta),{y}_{t,j}),\label{eq:loss}\end{equation}
where \(L_t(\cdot, \cdot)\) denotes the loss function for task \(T_t\), e.g., cross-entropy. 

We treat pretraining as an inaccessible stage that precedes \(T_1\).
When learning task \(T_t\), the model has no access to the original pretraining data, pretraining statistics, or earlier-task data.
We also keep the model size fixed across tasks to avoid accumulating task-specific modules and increasing storage or deployment overhead.

SVC follows the parameter-efficient fine-tuning paradigm and operates on linear layers.
Consider the \(l\)-th linear layer before learning task \(T_t\).
Its inherited weight matrix is
\(
\mathbf{W}_{t-1}^{(l)}
\in
\mathbb{R}^{
d_{\mathrm{out}}^{(l)}
\times
d_{\mathrm{in}}^{(l)}
}.
\)
Let
\(
\mathbf{X}_{t,\mathrm{in}}^{(l)}
\in
\mathbb{R}^{
N
\times
d_{\mathrm{in}}^{(l)}
}
\)
denote the input activations of this layer.
Here, \(N\) denotes the number of activation vectors.
The layer output is
\(\mathbf{X}_{t,\mathrm{out}}^{(l)}
=
\mathbf{X}_{t,\mathrm{in}}^{(l)}
\bigl(\mathbf{W}_{t-1}^{(l)}\bigr)^\top.\)
We decompose the weight matrix as 
\(
\mathbf{W}_{t-1}^{(l)}
=
\mathbf{U}_{t-1}^{(l)}
{\Sigma}_{t-1}^{(l)}
\bigl(\mathbf{V}_{t-1}^{(l)}\bigr)^\top
=
\sum_{i=1}^{r_l}
\sigma_{t-1,i}^{(l)}
\mathbf{u}_{t-1,i}^{(l)}
\mathbf{v}_{t-1,i}^{(l)\top},
\) where \(r_l=\operatorname{rank}(\mathbf{W}_{t-1}^{(l)})\). The corresponding linear transformation can therefore be written as
\(
    \mathbf{X}_{t,\mathrm{out}}^{(l)}
    =\sum_{i=1}^{r_l}
    \left(
    \mathbf{X}_{t,\mathrm{in}}^{(l)}
    \mathbf{v}_{t-1,i}^{(l)}
    \right)
    \sigma_{t-1,i}^{(l)}
    \mathbf{u}_{t-1,i}^{(l)\top},
\)
where \(\sigma_{t-1,i}^{(l)}\), \(\mathbf{u}_{t-1,i}^{(l)}\), and \(\mathbf{v}_{t-1,i}^{(l)}\) denote the \(i\)-th singular value, left singular vector, and right singular vector, respectively. 
The singular value controls the strength of the corresponding rank-one component, while the right and left singular vectors determine its input and output directions.
Motivated by the stable capability retention of singular-value fine-tuning, we treat the paired input and output directions associated with each singular value as a single adaptation unit.
We define the corresponding singular-vector channel as
~\footnote{Section~\ref{section:analyze} examines how including singular
values in each channel affects formulation and performance.}
$
\mathcal{S}_{t-1,i}^{(l)}
:=
\left(
\mathbf{u}_{t-1,i}^{(l)},
\mathbf{v}_{t-1,i}^{(l)}
\right).
$
SVC then selectively enables updates to these channels while keeping all singular values trainable.

\subsection{Benefit–Cost Estimation for Singular-Vector Channels}
\label{section 3.2}
We assess each singular-vector channel by its adaptation benefit and forgetting cost. Since task \(T_t\) is learned by minimizing the current-task loss in Equation~\ref{eq:loss}, the loss reduction resulting from a channel update provides a natural measure of its adaptation benefit.
Forgetting cost, however, cannot be evaluated in the same way, because the original historical data are unavailable and a unified loss is difficult to construct for capturing changes in the model’s broad general capabilities.
Inspired by methods such as GPM~\citep{saha2021gradient} that constrain updates based on activation subspaces, we therefore use linear-layer output drift on historical activations as a local proxy for forgetting cost.
Although these activations are unavailable, their effect can be approximated using activation statistics collected from unlabeled general-domain data.
Let \(\Delta\mathbf W_{t,i}^{(l)}\) denote the directional weight change induced by updating singular-vector channel \(\mathcal S_{t-1,i}^{(l)}\).
The corresponding adaptation benefit \(B_{t,i}^{(l)}\) and forgetting cost \(C_{t,i}^{(l)}\) are conceptually defined as:
\begin{equation}B_{t,i}^{(l)}=\mathcal L_t\left(\mathbf W_{t-1}^{(l)}\right)-\mathcal L_t\left(\mathbf W_{t-1}^{(l)}+\Delta\mathbf W_{t,i}^{(l)}\right),\end{equation}
\begin{equation}
C_{t,i}^{(l)}=\mathbb{E}_{x \sim \mathcal{D}_{\text{old}}}
\left\|\Delta\mathbf W_{t,i}^{(l)}x^{(l)}\right\|_2^2
=\operatorname{tr}\left(\Delta\mathbf W_{t,i}^{(l)} \Sigma_{\text{old}}^{(l)} \Delta\mathbf W_{t,i}^{(l)\top}\right),
\quad
\label{eq:cost_paper}
\end{equation}
where \(\Sigma_{\mathrm{old}}^{(l)}=\mathbb{E}_{x^{(l)}\sim\mathcal{D}_{\mathrm{old}}}[x^{(l)}x^{(l)\top}]\in\mathbb{R}^{
d_{\mathrm{in}}^{(l)}
\times
d_{\mathrm{in}}^{(l)}
}\) denotes the second-moment matrix of historical input activations at layer \(l\), serving here as an ideal reference statistic.

However, separately optimizing each channel to obtain \(\Delta\mathbf{W}_{t,i}^{(l)}\) would incur prohibitive computational overhead. 
We therefore use the average current-task gradient at layer \(l\) as the basis for estimating each channel's initial directional update tendency:
\begin{equation}
    \mathbf G_t^{(l)}
    =
    \frac{1}{n_t}
    \sum_{j=1}^{n_t}
    \nabla_{\mathbf W^{(l)}_{t-1}}
    L_{t}\left(
    f(\mathbf x_{t,j};{\theta}_{t-1}),
    \mathbf y_{t,j}
    \right),
\end{equation}
After omitting positive constants shared by all channels within each objective, including the learning-rate terms, the resulting estimates take the following form:
\begin{equation}
    B_{t,i}^{(l)}
    =
    \|\mathbf p_{t,i}^{(l)}\|_2^2
    +
    \|\mathbf q_{t,i}^{(l)}\|_2^2,
\label{eq:benfit_paper_final}
\end{equation}
\begin{equation}
    C_{t,i}^{(l)}
    =
    \bigl(\sigma_{t-1,i}^{(l)}\bigr)^2
    \left[
    \left(
    \mathbf v_{t-1,i}^{(l)\top}
    {\Sigma}_{\mathrm{old}}^{(l)}
    \mathbf v_{t-1,i}^{(l)}
    \right)
    \|\mathbf p_{t,i}^{(l)}\|_2^2
    +
    \mathbf q_{t,i}^{(l)\top}
    {\Sigma}_{\mathrm{old}}^{(l)}
    \mathbf q_{t,i}^{(l)}
    \right],
\label{eq:cost_paper_final}
\end{equation}
where \(\mathbf p_{t,i}^{(l)}\) and \(\mathbf q_{t,i}^{(l)}\) denote the effective gradients for updating the left and right singular-vector directions, respectively:
\begin{equation}
    \mathbf p_{t,i}^{(l)}
    =
    \left(
    \mathbf I-
    \mathbf u_{t-1,i}^{(l)}
    \mathbf u_{t-1,i}^{(l)\top}
    \right)
    \sigma_{t-1,i}^{(l)}
    \mathbf G_t^{(l)}
    \mathbf v_{t-1,i}^{(l)},
\end{equation}
\begin{equation}
    \mathbf q_{t,i}^{(l)}
    =
    \left(
    \mathbf I-
    \mathbf v_{t-1,i}^{(l)}
    \mathbf v_{t-1,i}^{(l)\top}
    \right)
    \sigma_{t-1,i}^{(l)}
    \mathbf G_t^{(l)\top}
    \mathbf u_{t-1,i}^{(l)}.
\end{equation}
These projected gradients are used only for channel scoring before fine-tuning and do not constrain the subsequent optimization of the selected singular-vector directions.

Intuitively, the adaptation benefit measures the magnitude of the initial directional update tendency.
The forgetting cost further weights this tendency by its interaction with the second-order statistics of historical activations.

Equations~\ref{eq:cost_paper} and~\ref{eq:cost_paper_final} show that an ideal evaluation of forgetting cost requires unavailable historical activations.
However, historical data enter the resulting score only through the second-moment matrix
\(\boldsymbol{\Sigma}_{\mathrm{old}}^{(l)}\),
whose estimation requires no ground-truth labels.
This allows us to approximate these statistics using a more accessible open-source, unlabeled general-domain corpus. In practice, we replace \(\boldsymbol{\Sigma}_{\mathrm{old}}^{(l)}\) with the second-moment matrix computed from the shared general-domain reference corpus.
Detailed derivations of the adaptation benefit and forgetting cost are provided in Appendix~A.

\subsection{Pareto-Frontier-Based Adaptive Channel Selection}
\label{section:3.3}
Given the scores estimated in Section~\ref{section 3.2}, we seek channels with low forgetting cost and sufficient adaptation benefit.
A weighted combination of the two objectives would require a trade-off coefficient that may vary across layers and tasks.
We therefore use a three-stage procedure: knee-based cost screening excludes high-cost channels, Pareto-front filtering removes dominated candidates, and an adaptive efficiency cutoff removes channels with unfavorable benefit–cost trade-offs.

First, we log-transform the scores to reduce their scale variation: \(b_{t,i}^{(l)}=\log(\max\{B_{t,i}^{(l)},\epsilon\})\) and \(c_{t,i}^{(l)}=\log(\max\{C_{t,i}^{(l)},\epsilon\})\), where \(\epsilon\) ensures numerical stability.
We sort the log-cost scores as
\(c_{t,(1)}^{(l)}\leq\cdots\leq c_{t,(r_l)}^{(l)}\) and apply Kneedle
to the resulting rank--cost curve, using a convex increasing configuration
with sensitivity \(S=1.0\)~\citep{satopaa2011finding}.
The detected knee \(k_t^{(l)}\) defines
\(c_{t,*}^{(l)}=c_{t,(k_t^{(l)})}^{(l)}\).
If no knee is detected, we set \(c_{t,*}^{(l)}=+\infty\), retaining all
channels at this stage.
The resulting candidate set is
\begin{equation}
\mathcal F_t^{(l)}
=
\left\{
\mathcal S_{t-1,i}^{(l)}
\;\middle|\;
c_{t,i}^{(l)}
\leq
c_{t,*}^{(l)}
\right\}.
\label{eq:cost_feasible_set}
\end{equation}

Within \(\mathcal F_t^{(l)}\), we retain the nondominated channels through Pareto-front filtering.
A channel \(\mathcal S_{t-1,i}^{(l)}\) dominates \(\mathcal S_{t-1,j}^{(l)}\) if it has no lower benefit and no higher cost, with at least one strict inequality.
The Pareto frontier is defined as
\begin{equation}
\label{eq:pareto_frontier}
\begin{aligned}
\mathcal{P}_t^{(l)}
&=
\left\{
\mathcal{S}_{t-1,j}^{(l)} \in \mathcal{F}_t^{(l)}
\,\middle|\,
\nexists\, \mathcal{S}_{t-1,i}^{(l)} \in \mathcal{F}_t^{(l)}
:
\mathcal{S}_{t-1,i}^{(l)}
\succ
\mathcal{S}_{t-1,j}^{(l)}
\right\},
\\
\mathcal{S}_{t-1,i}^{(l)}
\succ
\mathcal{S}_{t-1,j}^{(l)}
&\iff
\left(
b_{t,i}^{(l)} \geq b_{t,j}^{(l)}
\right)
\land
\left(
c_{t,i}^{(l)} \leq c_{t,j}^{(l)}
\right)
\\
&\qquad\land
\left[
\left(
b_{t,i}^{(l)} > b_{t,j}^{(l)}
\right)
\lor
\left(
c_{t,i}^{(l)} < c_{t,j}^{(l)}
\right)
\right].
\end{aligned}
\end{equation}

Pareto-front filtering removes dominated channels, but some remaining channels may still offer insufficient benefit relative to their costs. We therefore define the log-efficiency score
$
    r_{t,i}^{(l)}
    =
    b_{t,i}^{(l)}
    -
    c_{t,i}^{(l)}.
    \label{eq:benefit_cost_ratio}
$

Let
\(\mathcal I_{P,t}^{(l)}
=
\left\{
i\in\{1,\ldots,r_l\}
\;\middle|\;
\mathcal S_{t-1,i}^{(l)}
\in\mathcal P_t^{(l)}
\right\},\)
and \(\pi_t^{(l)}\) be a permutation of
\(\mathcal I_{P,t}^{(l)}\) that orders its channels by decreasing
log-efficiency:
\(r_{t,\pi_t^{(l)}(1)}^{(l)}
\geq
\cdots
\geq
r_{t,\pi_t^{(l)}(|\mathcal I_{P,t}^{(l)}|)}^{(l)}.\)
We apply Otsu's method to the efficiency scores of the Pareto-front
channels~\citep{otsu1979threshold}.
If it yields a valid two-group partition, channels with scores no smaller
than the threshold are selected.
Otherwise, all channels in \(\mathcal P_t^{(l)}\) are retained.
The resulting cutoff determines the number of selected channels
\(K_t^{(l)}\).
The final selected set is
\begin{equation}
\mathcal A_t^{(l)}
=
\left\{
\mathcal S_{t-1,\pi_t^{(l)}(k)}^{(l)}
\;\middle|\;
1\leq k\leq K_t^{(l)}
\right\}.
\label{eq:selection_final}
\end{equation}

\subsection{SVC}
We illustrate the SVC fine-tuning procedure on task \(T_t\).
Before fine-tuning, SVC computes layer-wise activation second moments from an unlabeled general-domain corpus and average gradients from the current-task data.
SVC then applies SVD to each layer and scores its singular-vector channels using Equations~\ref{eq:benfit_paper_final} and~\ref{eq:cost_paper_final}.
For each layer, SVC obtains \(\mathcal A_t^{(l)}\) through knee-based cost screening, Pareto-front filtering, and adaptive thresholding, as defined in Equations~\ref{eq:cost_feasible_set}--\ref{eq:selection_final}.
This selection is completed before fine-tuning and remains fixed throughout training on \(T_t\).

During fine-tuning, \(\mathbf W_{t-1}^{(l)}\) remains frozen, and only structured increments are optimized.
Specifically, all singular values are assigned trainable increments, whereas singular-vector increments are registered only for channels in \(\mathcal A_t^{(l)}\).
Let \(\delta\sigma_{t,i}^{(l)}\), \(\delta\mathbf u_{t,i}^{(l)}\), and \(\delta\mathbf v_{t,i}^{(l)}\) denote the corresponding increments, and let \(N_{\mathrm{lin}}\) denote the number of linear layers adapted by SVC.
The trainable parameter sets for layer \(l\) and task \(T_t\) are
\begin{equation}
    \Phi_t^{(l)}
    =
    \left\{\delta\sigma_{t,i}^{(l)}\right\}_{i=1}^{r_l}
    \cup
    \left\{
    (\delta\mathbf u_{t,i}^{(l)},
    \delta\mathbf v_{t,i}^{(l)})
    \;\middle|\;
    \mathcal S_{t-1,i}^{(l)}\in\mathcal A_t^{(l)}
    \right\},
    \qquad
    \Phi_t
    =
    \bigcup_{l=1}^{N_{\mathrm{lin}}}
    \Phi_t^{(l)}.
\end{equation}

All incremental parameters are initialized to zero. During fine-tuning, the updated factors and the forward pass of layer \(l\) are given by
\begin{equation}
\begin{aligned}
\widetilde{\sigma}_{t,i}^{(l)}
&=
\sigma_{t-1,i}^{(l)}
+
\delta\sigma_{t,i}^{(l)},\\
\left(
\widetilde{\mathbf u}_{t,i}^{(l)},
\widetilde{\mathbf v}_{t,i}^{(l)}
\right)
&=
\begin{cases}
\left(
\mathbf u_{t-1,i}^{(l)}+\delta\mathbf u_{t,i}^{(l)},
\mathbf v_{t-1,i}^{(l)}+\delta\mathbf v_{t,i}^{(l)}
\right),
&
\mathcal S_{t-1,i}^{(l)}\in\mathcal A_t^{(l)},\\
\left(
\mathbf u_{t-1,i}^{(l)},
\mathbf v_{t-1,i}^{(l)}
\right),
&
\mathcal S_{t-1,i}^{(l)}\notin\mathcal A_t^{(l)},
\end{cases}\\
\mathbf X_{t,\mathrm{out}}^{(l)}
&=
\sum_{i=1}^{r_l}
\left(
\mathbf X_{t,\mathrm{in}}^{(l)}
\widetilde{\mathbf v}_{t,i}^{(l)}
\right)
\widetilde{\sigma}_{t,i}^{(l)}
\widetilde{\mathbf u}_{t,i}^{(l)\top}.
\end{aligned}
\end{equation}
After fine-tuning, the learned factor changes are merged into the inherited weight:
\(    \mathbf W_t^{(l)}
    =
    \mathbf W_{t-1}^{(l)}
    +
    \sum_{i=1}^{r_l}
    \left[
    \widetilde{\sigma}_{t,i}^{(l),*}
    \widetilde{\mathbf u}_{t,i}^{(l),*}
    \widetilde{\mathbf v}_{t,i}^{(l),*\top}
    -
    \sigma_{t-1,i}^{(l)}
    \mathbf u_{t-1,i}^{(l)}
    \mathbf v_{t-1,i}^{(l)\top}
    \right],\)
where the superscript \(*\) denotes the corresponding values after fine-tuning. The merged weight \(\mathbf W_t^{(l)}\) is then carried forward to task \(T_{t+1}\). Thus, SVC requires no persistent task-specific modules and introduces no task-wise model growth. 
\section{Experiments}
\label{sec:4}
We evaluate SVC across model families and against continual learning baselines.
\subsection{Experimental Setup}
\label{section es}
\textbf{Datasets and Metrics.}
For sequential learning, we use TRACE~\citep{wang2023trace}, covering science (ScienceQA), finance (FOMC), summarization (MeetingBank), multilingual tasks (German 20Minuten and Chinese C-STANCE), code (Py150), and mathematics (NumGLUE-cm and NumGLUE-ds).
Classification and single-token question-answering tasks are evaluated by accuracy, while other generation tasks use the average of ROUGE-L and BLEU. 
Let \(R_{t,i}\) denote performance on task \(i\) after learning task \(t\). 
We report Overall Performance (OP)~\citep{chaudhry2018riemannian}, \(OP_t=\frac{1}{t}\sum_{i=1}^{t}R_{t,i}\), and Backward Transfer (BWT)~\citep{lopez2017gradient}, \(BWT_t=\frac{1}{t-1}\sum_{i=1}^{t-1}(R_{t,i}-R_{i,i})\), which measure average learned-task performance and subsequent changes on previous tasks, respectively.
We measure capability preservation using Pass@1 on HumanEval~\citep{chen2021evaluating} for code generation and GSM8K~\citep{cobbe2021training} for mathematical reasoning.
To estimate SVC's activation second moments, we sample 30,000 unlabeled examples from the open-source C4~\citep{raffel2020exploring} and Stack-v2-Python~\citep{lozhkov2024starcoder} datasets at a 2:1 ratio.
This corpus is fixed across all models and tasks.
Details are provided in Appendix~B.

\textbf{Baselines.}
We compare SVC with four baseline categories.
\textbf{Regularization} methods include EWC~\citep{kirkpatrick2017overcoming}, GEM~\citep{lopez2017gradient}, and LwF~\citep{li2017learning}, which preserve prior knowledge through parameter-importance constraints, gradient projection, and knowledge distillation, respectively. 
\textbf{Replay} reuses past data after each task (Replay)~\citep{rebuffi2017icarl} or throughout training (Replay-online)~\citep{rolnick2019experience}.
\textbf{PEFT} includes SeqLoRA, O-LoRA~\citep{wang2023orthogonal}, PiSSA~\citep{meng2024pissa}, and MiLoRA~\citep{wang2025milora}.
O-LoRA enforces orthogonality among task-specific LoRA subspaces.
PiSSA updates adapter components initialized from the largest singular values, whereas MiLoRA uses the smallest-value components while freezing the dominant ones.
PiSSA and MiLoRA thus make opposing fixed selections based on singular-value magnitude.
In addition, we include OSFT~\citep{nayak2026sculpting}, an SVD-based constrained full-parameter fine-tuning method.
OSFT uses previous-task data to identify important parameter subspaces and restricts new-task updates to their orthogonal complements. Sequential full-parameter fine-tuning (SeqFT) is also included. 
Details are provided in Appendix~C.

\begin{table*}[t]
\centering
\caption{Main results across four LLMs and three task orders. HumanEval and GSM8K measure general-capability preservation, while OP and BWT evaluate sequential task learning. Higher is better for all metrics. Best and second-best results are shown in bold and underlined, respectively.}
\label{tab1}
\resizebox{\textwidth}{!}{
\setlength{\tabcolsep}{1.4mm}{
\renewcommand{\arraystretch}{1.1}
\newcolumntype{C}{>{\centering\arraybackslash}p{0.96cm}}
\begin{tabular}{
    l
    !{\hspace{0.8mm}}
    *{4}{C}
    !{\hspace{0.8mm}}
    *{4}{C}
    !{\hspace{0.8mm}}
    *{4}{C}
}
\hlineB{4}
\multirow{2}{*}{Methods} & \multicolumn{4}{c}{Order 1} & \multicolumn{4}{c}{Order 2} & \multicolumn{4}{c}{Order 3} \\ 
\cmidrule(l{0pt}r{1.2em}){2-5}
\cmidrule(l{0pt}r{1.2em}){6-9}
\cmidrule(l{0pt}r{1.2em}){10-13}
& \makebox[0pt][c]{
  \scalebox{0.92}[1]{\small HumanEval}\hspace{1.2mm}
}
&
\makebox[0pt][c]{
  \hspace{1.2mm}\scalebox{0.94}[1]{\small GSM8K}
}
& \small OP
& \small BWT
& \makebox[0pt][c]{
  \scalebox{0.92}[1]{\small HumanEval}\hspace{1.2mm}
}
&
\makebox[0pt][c]{
  \hspace{1.2mm}\scalebox{0.94}[1]{\small GSM8K}
}
& \small OP
& \small BWT
& \makebox[0pt][c]{
  \scalebox{0.92}[1]{\small HumanEval}\hspace{1.2mm}
}
&
\makebox[0pt][c]{
  \hspace{1.2mm}\scalebox{0.94}[1]{\small GSM8K}
}
& \small OP
& \small BWT
\\ \hline
\multicolumn{13}{l}{
    \rule[-0.5ex]{0pt}{3.0ex}\emph{Llama3-8B}
} \\ \hline
SeqFT           & 0.0  & 2.88  & 39.22 & -10.69 & 0.61  &	4.32  & 29.78 & -10.25 & 0.0  & 3.11  & 20.10 & -28.79 \\
EWC             & 0.0  & 3.26  & 32.95 & -10.05 & 0.0  &	5.46  & 34.33 & -11.27 & 2.44 	& 3.26 &	36.92 &	-6.74 \\
GEM             & 0.0  & 3.34  & 39.37 & -10.24 & 0.0  &	2.58  & 32.03 & -12.44 & 0.0  & 2.81  & 39.07 & -7.15 \\
LwF             & 0.0  & 3.03  & 33.06 & -9.94  & 1.83  &	2.73  & 27.20 & -8.30  & 1.22  & 2.50  & 26.69 & -13.53 \\
Replay          & 0.0  & 2.05  & 38.62 & -2.94  & 0.61 &	3.26 &	41.90 &	\underline{-0.24} & 0.61 &	3.34 &	39.56 &	-3.87 \\
Replay-online   & 0.0  & 3.11  & 45.47 & -1.99  & 0.0  &	3.03  & 41.33 & -1.48  & 0.0  & 3.79  & 40.90 & \underline{-1.64}  \\
SeqLoRA         & 28.05 & 44.20 & 46.04 & -4.39  & 27.44 & 	21.83 & 44.36 &	-7.62  & \textbf{30.49} & 36.39 & 32.14 & -2.80 \\
O-LoRA          & 26.83 & 42.84 & 39.73 & -7.19 & 26.22 &	39.95 &	41.39 &	-4.07 &	29.27 &	{37.91} &	{43.43} &	-2.43  \\
PiSSA           & 28.05 &	39.42 &	47.81 &	-2.47 &	\underline{30.49} &	21.30 &	47.54 &	-2.33 &	28.05 &	16.45 &	\underline{47.21} &	-2.92 \\
MiLoRA          & \underline{29.27} & \underline{45.49} & \underline{47.90} & \underline{-1.55} & 28.66 &	\underline{41.24} &	\underline{47.66} &	-2.38 &	\textbf{30.49} &	\underline{46.17} &	44.15 &	-4.21 \\
OSFT            & 0.0  & 1.97  & 28.00 & -11.82 & 0.61  &	2.65 &	31.38 &	-7.47  & 0.0  & 2.96  & 27.37 & -13.55 \\
\rowcolor{black!6}
SVC (ours)    & \textbf{30.49} & \textbf{51.55} & \textbf{49.62} & \textbf{-0.94}  & \textbf{31.10} &	\textbf{51.02} &	\textbf{50.48} &	\textbf{-0.05}  & \underline{29.88} & \textbf{47.31} & \textbf{49.21} & \textbf{-1.43} \\
\hline
\multicolumn{13}{l}{
    \rule[-0.5ex]{0pt}{3.0ex}\emph{Mistral-7B-v0.3}
} \\ \hline
SeqFT           & 0.0 &	1.14 &	14.76 &	-22.20 &	0.0 &	1.97 &	23.79 &	-12.85 &	0.0 &	0.99 &	13.20 &	-23.57 \\
EWC             & 0.0 &	1.06 &	17.65 &	-18.72 &	0.0 &	2.05 &	22.44 &	-13.20 &	0.0 &	1.90 &	15.11 &	-22.43 \\
GEM             & 0.0 &	2.05 &	16.80 &	-21.40 & 	0.0 &	2.27 &	26.85 &	-8.73 &	0.0 &	0.68 &	15.13 &	-21.82 \\
LwF             & 0.0 &	0.0 &	7.13 &	-23.47 &	0.0 &	0.0 &	16.37 &	-9.83 &	0.0 &	0.76 &	12.40 &	-18.06 \\
Replay          & 0.0 &	0.45 &	17.11 &	{-5.10} &	0.0 &	1.97 &	30.14 &	\textbf{-1.06} &	0.0 &	2.12 &	31.30 &	-2.37  \\
Replay-online   & 0.0 &	0.0 &	17.36 &	-6.46 &	0.0 &	1.52 &	32.66 &	\underline{-1.28} &	0.0 &	1.52 &	33.17 &	\textbf{-1.02}  \\
SeqLoRA         & {16.46} &	{29.11} &	45.67 &	-3.91 &	15.24 &	20.85 &	41.63 &	-11.31 &	\underline{19.51} &	{12.96} &	\textbf{45.17} & -5.48  \\
O-LoRA          & 12.20 & 24.64 & {40.29} & -5.16 & {17.07} & {23.35} &	{35.92} &	-11.31 & 10.37 &	12.74 &	{39.36} &	-5.38 \\
PiSSA           & 21.34 &	11.90 &	38.98 &	-3.70 &	16.46 &	19.86 &	\underline{44.67} &	-3.50 &	11.59 &	3.79 &	44.26 &	-1.73 \\
MiLoRA          & \underline{25.61} &	\underline{30.78} &	\underline{46.44} &	\underline{-2.56} &	\underline{24.39} &	\underline{26.76} &	42.47 &	-5.05 &	\textbf{25.61} &	\underline{28.05} &	43.45 &	-2.75 \\
OSFT            & 0.0 &	2.43 &	13.36 &	-18.42 &	0.0 &	0.0 &	2.20 &	-19.86 &	0.0 &	0.0 &	2.54 &	-16.52 \\
\rowcolor{black!6}
SVC (ours)    & \textbf{27.44} &	\textbf{36.54} &	\textbf{48.37} &	\textbf{-0.37} &	\textbf{29.88} &	\textbf{37.07} &	\textbf{48.70} &	-1.75 &	\textbf{25.61} &	\textbf{37.68} &	\underline{44.78} &	\underline{-1.53} \\
\hline
\multicolumn{13}{l}{
    \rule[-0.5ex]{0pt}{3.0ex}\emph{Gemma2-9B-it}
} \\ \hline
SeqFT           & 9.76 &	5.38 &	34.09 &	-12.37 &	7.32 &	5.53 &	34.77 &	-7.30 &	7.32 &	1.74 &	33.97 &	-11.17 \\
EWC             & 7.32 &	2.73 &	38.58 &	-6.96 &	5.49 &	6.75 &	36.85 &	-8.90 &	9.15 &	2.12 &	38.72 &	-4.53  \\
GEM             & 10.37 & 	3.11 &	32.55 &	-8.98 &	10.98 &	6.67 &	37.22 &	-5.70 &	5.49 &	1.97 &	33.89 &	-13.20 \\
LwF             & 0.0 &	2.50 &	26.51 &	-12.52 &	1.22 &	2.58 &	23.76 &	-10.12 &	1.22 &	3.87 &	28.18 &	-9.94  \\
Replay          & 6.71 &	3.26 &	42.30 &	\textbf{-0.66} &	0.61 &	3.49 &	39.44 &	\textbf{0.51} &	3.05 &	2.35 &	41.78 &	{-1.32}   \\
Replay-online   & 3.66 &	2.58 &	44.12 &	{-1.95} &	1.22 &	2.73 &	38.61 &	-3.30 &	3.05 &	1.97 &	40.25 & -3.32   \\
SeqLoRA         & 37.20 & 	{73.09} &	50.19 &	-6.83 &	32.32 &	56.86 &	{51.58} &	-4.63 &	{28.66} &	3.94 &	51.97 &	-4.56 \\
O-LoRA          & {41.46} &	63.08 &	{47.91} &	-5.88 &	{37.80} &	{57.01} &	45.43 &	-2.60 & 15.85 &	{46.40} &	{46.00} &	-7.03 \\
PiSSA           & 49.39 &	67.63 &	\textbf{51.93} &	-3.60 &	\underline{48.78} &	66.19 &	\underline{53.90} &	-0.92 &	\underline{48.78} &	\underline{55.04} &	52.45 &	-2.95 \\
MiLoRA          & \underline{50.61} &	\underline{73.54} &	{49.91} &	-4.21 &	47.56 &	\underline{73.69} &	53.20 &	-1.54 &	44.51 &	5.53 &	\underline{54.31} &	\underline{-0.91} \\
OSFT            & 7.32 &	3.41 &	30.35 &	-4.29 &	0.0 &	0.99 &	19.14 &	-12.29 &	0.0 &	0.76 &	11.14 &	-23.95  \\
\rowcolor{black!6}
SVC (ours)    & \textbf{56.10} &	\textbf{75.51} &	\underline{50.25} &	\underline{-1.37} &	\textbf{53.66} &	\textbf{74.22} &	\textbf{54.95} &	\underline{-0.72} &	\textbf{53.66} &	\textbf{74.91} &	\textbf{55.84} &	\textbf{-0.11} \\
\hline
\multicolumn{13}{l}{
    \rule[-0.5ex]{0pt}{3.0ex}\emph{Qwen3.5-9B-Base}
} \\ \hline
SeqFT           & 20.73 &	11.07 &	50.23 &	-3.66 &	23.78 &	49.58 &	46.18 &	-7.47 &	17.68 &	4.25 &	48.51 &	-5.24 \\
EWC             & 21.95 &	11.75 &	47.88 &	-5.57 &	20.12 &	48.07 &	47.38 &	-3.60 &	16.46 &	4.70 &	48.01 &	-3.59 \\
GEM             & 18.29 &	14.40 &	49.64 &	-3.38 &	16.46 &	41.62 &	45.15 &	-7.32 &	18.29 &	4.25 &	47.26 &	-3.65 \\
LwF             & 16.46 &	28.73 &	38.23 &	-8.44 &	21.34 &	23.73 &	31.94 &	-9.66 &	22.56 &	30.17 &	38.27 &	-9.17 \\
Replay          & 29.88 &	16.07 &	49.36 &	-0.99 &	17.07 &	12.89 &	49.97 &	-0.94 &	12.20 &	5.23 &	46.70 &	{-1.04} \\
Replay-online   & 24.39 &	12.13 &	{51.16} &	{-0.78} &	28.66 &	36.39 &	\underline{51.91} &	\textbf{-0.16} &	12.20 &	12.05 &	47.25 &	-1.79 \\
SeqLoRA         & \underline{50.61} &	{71.19} &	49.86 &	-4.58 & \underline{50.61} &	{66.94} &	43.09 &	-7.46 &	{49.39} &	\textbf{74.60} &	{50.39} &	-4.64 \\
O-LoRA          & 28.05 & 42.99 & 39.11 & -3.80 & 34.76 &	39.12 &	42.01 &	-6.88 &	23.17 & 27.82 &	45.05 &	-6.74 \\
PiSSA           & \textbf{54.27} & \underline{71.87} & 51.46 & -1.31 &	48.17 &	67.17 &	48.09 &	-7.19 &	\underline{51.22} &	68.92 &	\underline{54.56} &	0.13 \\
MiLoRA          & 42.07 & 68.92 &	\underline{54.34} &	\textbf{0.59} &	49.39 &	\underline{70.13} &	44.71 &	-10.09 &	47.56 &	69.22 &	54.28 &	\textbf{2.36} \\
OSFT            & 11.59 &	5.31 &	47.99 &	-2.04 &	17.68 &	48.14 &	49.04 &	-1.78 &	1.22 &	4.55 &	45.83 &	-5.97 \\
\rowcolor{black!6}
SVC (ours)    & {43.29} &	\textbf{72.25} &	\textbf{57.05} &	\underline{0.02} &	\textbf{56.10} &	\textbf{71.27} &	\textbf{56.65} &	\underline{-0.53} &	\textbf{53.05} &	\underline{70.20} &	\textbf{57.87} &	\underline{0.76}  \\
\hlineB{4}
\end{tabular}}}
\vspace{-12pt}
\end{table*}

\textbf{Models and Implementation.}
We evaluate SVC on Llama3-8B~\citep{grattafiori2024llama}, Mistral-7B-v0.3~\citep{jiang2023mistral7b}, Gemma2-9B-it~\citep{team2024gemma}, and Qwen3.5-9B-Base~\citep{qwen35blog}, covering four representative open-source model families. 
All methods use Adam, a batch size of 64, no warmup, and five epochs per task.
Learning rates are \(10^{-4}\) for SVC and all PEFT baselines and \(10^{-5}\) otherwise.
Given the substantial impact of task order on continual learning~\citep{Yoon2020Scalable}, drawing on the experimental setups of~\citet{he2024seekr} and~\citet{wang2025parameter}, we evaluate all methods under three task orders. All experiments run on four 80-GB NVIDIA A100 GPUs. 
Further experimental details are provided in Appendix~C.

\subsection{Main Results}
Table~\ref{tab1} reports general-capability preservation and sequential task learning across four models and three task orders. We draw the following observations:\\
\textbf{(1) SVC effectively preserves general capabilities.} 
Across the 12 model and task-order settings, SVC achieves or ties the best result in 10 HumanEval settings and 11 GSM8K settings.
Against the strongest baseline in each setting, SVC achieves mean gains of 1.68 percentage points on HumanEval and 5.18 percentage points on GSM8K. Together, these results show that SVC provides the strongest overall preservation of general capabilities across model families and task orders. \\
\textbf{(2) SVC achieves a strong stability–plasticity balance.}
SVC obtains the best OP in nearly all settings across the four models and three task orders, while its BWT results also indicate substantially reduced forgetting. Despite storing no previous-task samples, SVC consistently surpasses both replay baselines in OP and remains competitive with them in BWT. Together, these results show that SVC acquires new-task knowledge while limiting interference with previously learned tasks.\\
\textbf{(3) SVC is robust across task orders.} 
Although task order affects final performance, SVC maintains strong general-capability preservation and sequential task learning across all three orders.
It ranks first or second in 46 of the 48 combinations of model, task order, and metric.
These results demonstrate that SVC remains robust to task order across different model families.

\subsection{Analysis and Discussion}
\label{section:analyze}
To understand how SVC balances stability and plasticity, we examine three questions about its channel-level design on Llama3-8B under Order 1.

\begin{table*}[t]
    \centering
    \vspace{-5pt}
    \caption{
    Validation of channel-level benefit and cost scores on Llama3-8B under
Order 1.
    }
    \vspace{-5pt}
    \label{tab:score_validation}
    \begingroup
    \footnotesize
    \setlength{\tabcolsep}{3pt}
    \renewcommand{\arraystretch}{0.90}
    
    \begin{tabular}{
        @{}lccc@{\hspace{11pt}}cccc@{}
    }
        \toprule
        Metric
        & High \(C\)
        & Mid \(C\)
        & Low \(C\)
        & Low \(B\), Mid \(C\)
        & High \(B\), Mid \(C\)
        & Low \(B\), Low \(C\)
        & High \(B\), Low \(C\) \\
        \midrule
        HumanEval
        & 0.00 & 28.05 &  {29.27}
        & 27.44 & 28.66 & 29.88 & 29.88 \\
        GSM8K
        & 2.65 & 44.05 &  {51.40}
        & 45.49 & 44.35 & 52.08 & 51.71 \\
        OP
        & 29.27 & 47.99 &  {51.12}
        & 47.26 & 47.63 & 50.66 & 50.28 \\
        BWT
        & -8.32 & -1.06 &  {-0.20}
        & -2.21 & -2.51 & 0.05 & -1.19 \\
        AP
        & 36.55 & 48.92 &  {51.30}
        & 49.19 &  {49.83}
        & 50.62 &  {51.32} \\
        \bottomrule
    \end{tabular}
        \vspace{-15pt}
    \endgroup
\end{table*}

\noindent\textbf{\emph{1. Are benefit and cost scores meaningful?}}
Because SVC selects singular-vector channels to balance learning and forgetting, we test whether the scores defined in Section~\ref{section 3.2} reflect actual update outcomes.
We first test whether \(C\) identifies risky update directions.
For each task and linear layer, 
we match SVC's channel count and select channels from the highest-, middle-, or lowest-cost region, with all other settings fixed.
As shown in Table~\ref{tab:score_validation}, the retention of pretrained capabilities improves consistently as the cost of the selected channels decreases.
High-cost selection also affects sequential learning, suggesting that excessive cost can offset adaptation gains.
These results show that \(C\) distinguishes risky channel updates and provides a reliable basis for controlling forgetting.

To reduce the influence of \(C\) when assessing \(B\), we compare channels under cost-matched conditions. 
For each task and layer, we rank channels by \(C\) and separately take \(2K\) candidates from the lowest- and middle-cost regions, where \(K\) is SVC's channel count.
We pair cost-adjacent channels and select the higher- or lower-\(B\) channel from each pair.
Within each region, this produces two disjoint \(K\)-channel sets with closely matched costs.
Because \(B\) measures current-task adaptation, we report
\(AP_t=\frac{1}{t}\sum_{i=1}^{t}R_{i,i}\), which averages performance immediately after each task is learned.
As shown in Table~\ref{tab:score_validation}, the high-\(B\) groups achieve higher AP than their low-\(B\) counterparts in both cost regions. 
Although strict cost matching limits the magnitude of the differences between the two groups, the trend remains consistent across both regions. This consistent trend indicates that \(B\) captures the contribution of a channel to current-task adaptation. Together, the results for \(B\) and \(C\) support their use in balancing learning and forgetting at the level of singular-vector channels.

\begin{wrapfigure}{r}{0.44\textwidth}
    \vspace{-20pt}
    \includegraphics[
        width=\linewidth
    ]{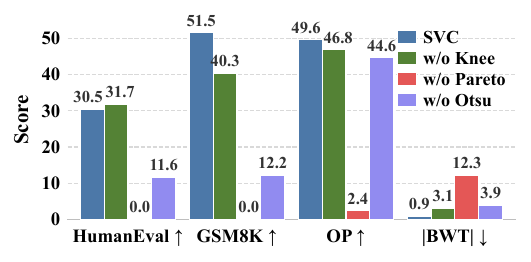}
    \vspace{-20pt}
    \caption{Ablation of the channel-selection stages. BWT is reported in absolute value.}
    \label{fig:result1}
    \vspace{-5pt}
\end{wrapfigure}
\noindent\textbf{\emph{2. How important is each selection stage?}}
Section~\ref{section:3.3} combines knee-based cost screening, Pareto-front filtering, and Otsu thresholding of benefit--cost efficiency scores.
We remove each stage in turn, keep all other settings fixed, and let the remaining stages determine the channel count.
Figure~\ref{fig:result1} shows that removing Pareto-front filtering causes the largest degradation, nearly eliminating the retention of code-generation and mathematical-reasoning capabilities and sharply reducing sequential performance.
Removing knee-based cost screening has a smaller effect but still weakens mathematical-reasoning retention, lowers sequential performance, and increases forgetting.
These results show that reliable channel selection depends on the complete three-stage procedure. Pareto filtering forms the core, while knee-based screening and the Otsu-based cutoff further restrict high-cost and low-efficiency channels to maintain the overall stability-plasticity balance.

\noindent\textbf{\emph{3. Should singular values be selected?}}
By default, SVC selectively updates singular vectors while training all singular values.
To test whether singular values should also be selected, we redefine each channel as \((\sigma_i,u_i,v_i)\), recompute its benefit and cost from the joint perturbation, and apply the same three-stage selection procedure. 
See Appendix~A for the derivation.
Compared with default SVC, 
joint selection decreases HumanEval, GSM8K, OP, and BWT by 1.22, 2.80, 0.65, and 0.56 points, respectively.
One possible explanation is that including the singular-value term changes the score distributions, which may reduce the separation between channels and alter both the selected set and its size. Moreover, singular values continue to adjust throughout training, so a local estimate computed before training may not accurately reflect their final contribution. These results support the default SVC design, which selectively updates singular vectors while optimizing all singular values to better balance learning and forgetting.
\section{Conclusion}
We present Selective Singular-Vector Channel Adaptation (SVC), which treats singular-vector channels as the basic units for balancing adaptation benefits against forgetting costs in continual learning for large language models. SVC estimates both quantities using current-task gradients and activation statistics from an unlabeled general-domain reference corpus, and adaptively selects trainable channels accordingly. 
Experiments across four LLM families and three task orders show that SVC improves continual-learning performance while preserving pretrained general capabilities, offering a task-adaptive route to stable and scalable continual domain adaptation for LLMs.
In future work, we will investigate reference-corpus construction methods that provide broader capability coverage and greater robustness to distribution shifts, as well as channel-selection strategies that account for interactions among channels and adjust dynamically throughout training. We will further evaluate SVC over longer task sequences and a broader range of general-capability benchmarks to characterize the limits of its robustness and computational efficiency.

\bibliography{iclr2027_conference}
\bibliographystyle{iclr2027_conference}

\appendix
\section*{Appendix}
\section{Derivation of Adaptation Benefit and Forgetting Cost}
\label{app:A}
For notational simplicity, we suppress the task index \(t\) and layer index \(l\) throughout this subsection, while retaining the channel index \(i\). Accordingly, \(W\), \(G_{\mathrm{new}}\), \(\mathcal L_{\mathrm{new}}\), and \(\Sigma_{\mathrm{old}}\) correspond to \(\mathbf W_{t-1}^{(l)}\), \(\mathbf G_t^{(l)}\), \(\mathcal L_t\), and \(\boldsymbol\Sigma_{\mathrm{old}}^{(l)}\), respectively. The task and layer indices are restored directly in the main-text expressions.
Consider a linear layer \(y=Wx\), where \(W\in\mathbb{R}^{d_{\mathrm{out}}\times d_{\mathrm{in}}}\) is the weight matrix before fine-tuning and \(x\in\mathbb{R}^{d_{\mathrm{in}}}\) is the input activation. The SVD of \(W\) is
\begin{equation}
W = U\Sigma V^\top = \sum_{i=1}^{r}{\sigma}_i \mathbf{u}_i \mathbf{v}_i^\top,
\label{eq:SVD}
\end{equation}
where \(r=\mathrm{rank}(W)\). We refer to \(\mathcal S_i=(\mathbf{u}_i,\mathbf{v}_i)\) as the \(i\)-th singular-vector channel and to \(\sigma_i\) as its associated singular value. Together, they define the rank-one component \(\sigma_i\mathbf{u}_i\mathbf{v}_i^\top\). The following benefit and cost analysis is carried out at the granularity of individual singular-vector channels.

The adaptation benefit is naturally characterized by the reduction in domain loss. Forgetting, however, cannot be fully characterized by a single loss, so following methods such as GPM, we instead measure it by the output drift of the linear layer on old-task input activations. Let \(\mathcal{L}_{\mathrm{new}}(W)\) denote the new-task loss, and let \(\Sigma_{\mathrm{old}}=\mathbb{E}_{x\sim\mathcal{D}_{\mathrm{old}}}[xx^\top]\in\mathbb{R}^{d_{\mathrm{in}}\times d_{\mathrm{in}}}\) denote the second-moment matrix of old-task input activations. For the local update \(\Delta W_i\) associated with the \(i\)-th singular-vector channel, we define its adaptation benefit \(B_i\) and forgetting cost \(C_i\) as

\begin{equation}
B_i=\mathcal{L}_{\text{new}}(W)-\mathcal{L}_{\text{new}}(W+\Delta W_i)
\label{eq:benefit}
\end{equation}

\begin{equation}
C_i=\mathbb{E}_{x \sim \mathcal{D}_{\text{old}}}
[\left\|\Delta{W}_ix\right\|_2^2]
=\operatorname{tr}\left(\Delta W_i \Sigma_{\text{old}} \Delta W_i^\top\right)
\quad
\label{eq:cost}
\end{equation}

In the default SVC formulation, channel selection determines whether the two singular-vector directions \((\mathbf u_i,\mathbf v_i)\) are trainable, whereas all singular values are optimized regardless of the selection result. We therefore first derive the default benefit and cost scores from the singular-vector perturbations. We then include the singular-value perturbation separately to obtain the alternative scores used in the corresponding ablation.



For completeness, consider small perturbations of the associated singular value and the two singular vectors:
\[\sigma_i \to \sigma_i + \delta\sigma_i\, \qquad   \mathbf{u}_i \to \mathbf{u}_i + \delta\mathbf{u}_i, \qquad \mathbf{v}_i \to \mathbf{v}_i + \delta\mathbf{v}_i\]
The local weight update is
\begin{align}
\Delta W_i 
&= \delta \sigma_i \mathbf{u}_i \mathbf{v}_i^\top + \sigma_i \delta\mathbf{u}_i \mathbf{v}_i^\top + \sigma_i \mathbf{u}_i \delta\mathbf{v}_i^\top + O(\delta^2)  \nonumber\\
&= (\delta \sigma_i + \sigma_i\mathbf{u}_i^\top \delta \mathbf{u}_i + \sigma_i\delta \mathbf{v}_i^\top \mathbf{v}_i) \mathbf{u}_i \mathbf{v}_i^\top + \sigma_i \delta\mathbf{u}_{i, \perp} \mathbf{v}_i^\top + \sigma_i \mathbf{u}_i \delta\mathbf{v}_{i, \perp}^\top + O(\delta^2) \nonumber \\
&= \Delta\sigma_i\mathbf{u}_i\mathbf{v}_i^\top + \sigma_i \delta\mathbf{u}_{i, \perp} \mathbf{v}_i^\top + \sigma_i \mathbf{u}_i \delta\mathbf{v}_{i, \perp}^\top + O(\delta^2)
\label{eq:delta1}
\end{align}
where \(\Delta\sigma_i=\delta\sigma_i+\sigma_i\mathbf u_i^\top\delta\mathbf u_i+\sigma_i\delta\mathbf v_i^\top\mathbf v_i\), \(\mathbf{u}_i^\top\delta\mathbf{u}_{i,\perp}=0\), and \(\mathbf{v}_i^\top\delta\mathbf{v}_{i,\perp}=0\). Ignoring second-order perturbation terms, the default channel score isolates the orthogonal singular-vector perturbations. The singular-value perturbation is considered separately in the ablation derived below.

\begin{equation}
\Delta W_i^{\perp}=
\sigma_i \delta\mathbf{u}_{i,\perp}\mathbf{v}_i^\top
+\sigma_i \mathbf{u}_i\delta\mathbf{v}_{i,\perp}^\top .
\label{eq:deltaW_perp}
\end{equation}

Let \(G_{\mathrm{new}}=\nabla_W\mathcal{L}_{\mathrm{new}}(W)\) denote the gradient of the new-task loss with respect to the weight matrix. A first-order Taylor expansion of \(\mathcal{L}_{\mathrm{new}}\) at \(W\) gives
\begin{equation}
\mathcal{L}_{\mathrm{new}}(W+\Delta W_i^{\perp})-\mathcal{L}_{\mathrm{new}}(W)
\approx \left\langle G_{\mathrm{new}},\Delta W_i^{\perp}
\right\rangle.
\label{eq:taylor_loss}
\end{equation}
Expanding the inner product with~\eqref{eq:deltaW_perp} substituted in,
\begin{align}
\left\langle G_{\mathrm{new}},\Delta W_i^{\perp}
\right\rangle
&=\left\langle G_{\mathrm{new}},\sigma_i \delta\mathbf{u}_{i,\perp}\mathbf{v}_i^\top+\sigma_i \mathbf{u}_i\delta\mathbf{v}_{i,\perp}^\top\right\rangle \nonumber \\
&=\sigma_i\operatorname{tr}
\left(G_{\mathrm{new}}^\top\delta\mathbf{u}_{i,\perp}\mathbf{v}_i^\top\right)+\sigma_i\operatorname{tr}\left(G_{\mathrm{new}}^\top\mathbf{u}_i\delta\mathbf{v}_{i,\perp}^\top\right)\nonumber \\
&=\sigma_i\operatorname{tr}\left(\mathbf{v}_i^\top G_{\mathrm{new}}^\top\delta\mathbf{u}_{i,\perp}\right)+\sigma_i\operatorname{tr}
\left(\delta\mathbf{v}_{i,\perp}^\top G_{\mathrm{new}}^\top\mathbf{u}_i
\right)\nonumber \\
&=\left\langle\sigma_i G_{\mathrm{new}}\mathbf{v}_i,\delta\mathbf{u}_{i,\perp}\right\rangle+\left\langle \sigma_iG_{\mathrm{new}}^\top\mathbf{u}_i,\delta\mathbf{v}_{i,\perp}\right\rangle .
\label{eq:loss_decompose}
\end{align}
This identifies the gradients associated with \(\delta\mathbf{u}_{i,\perp}\) and \(\delta\mathbf{v}_{i,\perp}\):
\begin{equation}
{\mathbf{g}}_{u_i}=\sigma_i G_{\mathrm{new}}\mathbf{v}_i,
\qquad
{\mathbf{g}}_{v_i}=\sigma_i G_{\mathrm{new}}^\top\mathbf{u}_i .
\label{eq:unconstrained_gradient}
\end{equation}
Since \(\delta\mathbf u_{i,\perp}\) and \(\delta\mathbf v_{i,\perp}\) must satisfy \(\mathbf u_i^\top\delta\mathbf u_{i,\perp}=0\) and \(\mathbf v_i^\top\delta\mathbf v_{i,\perp}=0\), we project \({\mathbf g}_{u_i},{\mathbf g}_{v_i}\) onto the orthogonal complements of \(\mathbf u_i,\mathbf v_i\) to obtain the feasible effective gradients:
\begin{equation}
\mathbf{p}_{i}=(I-\mathbf{u}_i\mathbf{u}_i^\top){\mathbf{g}}_{u_i},
\qquad
\mathbf{q}_{i}=(I-\mathbf{v}_i\mathbf{v}_i^\top){\mathbf{g}}_{v_i}.
\label{eq:project_gradient_general}
\end{equation}
Updating along \(-\mathbf p_i,-\mathbf q_i\) with step size \(\eta\) and substituting into~\eqref{eq:deltaW_perp} yields the local weight update along the orthogonal perturbation directions:
\begin{equation}
\Delta W_i^{\perp}= -\eta \sigma_i \mathbf{p}_i\mathbf{v}_i^\top -\eta \sigma_i \mathbf{u}_i\mathbf{q}_i^\top 
\label{eq:orthogonal_update}
\end{equation}
Substituting~\eqref{eq:orthogonal_update} into the benefit definition,~\eqref{eq:benefit}, and using \(\mathbf u_i^\top\mathbf p_i=0\), \(\mathbf v_i^\top\mathbf q_i=0\) to eliminate the cross terms
\begin{align}
B_i
&=\mathcal{L}_{\mathrm{new}}(W)-\mathcal{L}_{\mathrm{new}}(W+\Delta W_i^{\perp})\nonumber \\
&\approx-\left\langle
G_{\mathrm{new}},\Delta W_i^{\perp}\right\rangle\nonumber \\
&=\eta\sigma_i\left\langle
G_{\mathrm{new}},\mathbf{p}_i\mathbf{v}_i^\top+\mathbf{u}_i\mathbf{q}_i^\top\right\rangle\nonumber \\
&=\eta\sigma_i \operatorname{tr}(G_{\text{new}}^\top \mathbf{p}_i \mathbf{v}_i^\top) + \eta\sigma_i \operatorname{tr}(G_{\text{new}}^\top \mathbf{u}_i \mathbf{q}_i^\top)\nonumber \\
&=\eta \operatorname{tr}((\sigma_i G_{\text{new}} \mathbf{v}_i)^\top \mathbf{p}_i ) + \eta \operatorname{tr}(\mathbf{q}_i^\top (\sigma_i G_{\text{new}}^\top \mathbf{u}_i))\nonumber \\
&=\eta \operatorname{tr}((\mathbf{p}_i + \sigma_i(\mathbf{u}_i^\top G_{\text{new}} \mathbf{v}_i)\mathbf{u}_i)^\top \mathbf{p}_i ) + \eta \operatorname{tr}(\mathbf{q}_i^\top (\mathbf{q}_i + \sigma_i (\mathbf{v}_i^\top G_{\text{new}}^\top \mathbf{u}_i)\mathbf{v}_i))\nonumber \\
&=\eta
\left(\|\mathbf{p}_i\|_2^2
+
\|\mathbf{q}_i\|_2^2\right).
\label{eq:benefit_perp_expand}
\end{align}
Similarly, substituting~\eqref{eq:orthogonal_update} into the cost definition,~\eqref{eq:cost},
\begin{align}
C_i
&=\operatorname{tr}\left(\Delta W_i^{\perp}\Sigma_{\mathrm{old}}(\Delta W_i^{\perp})^\top\right)\nonumber \\
&=\eta^2\sigma_i^2\operatorname{tr}\left[\left(\mathbf{p}_i\mathbf{v}_i^\top+\mathbf{u}_i\mathbf{q}_i^\top\right)\Sigma_{\mathrm{old}}\left(\mathbf{v}_i\mathbf{p}_i^\top+\mathbf{q}_i\mathbf{u}_i^\top\right)\right]\nonumber \\
&=\eta^2\sigma_i^2\operatorname{tr}\left(\mathbf{p}_i\mathbf{v}_i^\top\Sigma_{\mathrm{old}}\mathbf{v}_i\mathbf{p}_i^\top\right)+\eta^2\sigma_i^2\operatorname{tr}\left(\mathbf{p}_i\mathbf{v}_i^\top\Sigma_{\mathrm{old}}\mathbf{q}_i\mathbf{u}_i^\top\right)\nonumber \\
&\quad+
\eta^2\sigma_i^2\operatorname{tr}
\left(
\mathbf{u}_i\mathbf{q}_i^\top
\Sigma_{\mathrm{old}}
\mathbf{v}_i\mathbf{p}_i^\top
\right) 
+
\eta^2\sigma_i^2
\operatorname{tr}
\left(
\mathbf{u}_i\mathbf{q}_i^\top
\Sigma_{\mathrm{old}}
\mathbf{q}_i\mathbf{u}_i^\top
\right)
\nonumber \\
&=
\eta^2\sigma_i^2
\left[
(\mathbf{v}_i^\top\Sigma_{\mathrm{old}}\mathbf{v}_i)
\operatorname{tr}(\mathbf{p}_i\mathbf{p}_i^\top)
+
(\mathbf{v}_i^\top\Sigma_{\mathrm{old}}\mathbf{q}_i)
\operatorname{tr}(\mathbf{u}_i^\top\mathbf{p}_i)
\right.
\nonumber \\
&\quad\left.
+
(\mathbf{q}_i^\top\Sigma_{\mathrm{old}}\mathbf{v}_i)
\operatorname{tr}(\mathbf{p}_i^\top\mathbf{u}_i)
+
(\mathbf{q}_i^\top\Sigma_{\mathrm{old}}\mathbf{q}_i)
\operatorname{tr}(\mathbf{u}_i^\top\mathbf{u}_i)
\right]
\nonumber \\
&=
\eta^2\sigma_i^2
\left[
(\mathbf{v}_i^\top\Sigma_{\mathrm{old}}\mathbf{v}_i)
\|\mathbf{p}_i\|_2^2
+
\mathbf{q}_i^\top\Sigma_{\mathrm{old}}\mathbf{q}_i
\right].
\label{eq:cost_perp_expand}
\end{align}

After omitting the positive learning-rate factors shared by all channels,~\eqref{eq:benefit_perp_expand} and~\eqref{eq:cost_perp_expand} yield the default benefit and forgetting-cost scores reported in the main text.

\paragraph{Singular-value-inclusive ablation.}
For the ablation in which the channel is instead defined as \(\widetilde{\mathcal S}_i=(\sigma_i,\mathbf u_i,\mathbf v_i)\), we include the singular-value perturbation in the same local analysis.

Incorporating the singular-value perturbation, gradient descent gives
\begin{equation}
\Delta\sigma_i = -\eta g_i, \qquad g_i \triangleq \mathbf{u}_i^\top G_{\mathrm{new}}\mathbf{v}_i.
\label{eq:sigma_update}
\end{equation}
The full update of channel \(i\) is
\begin{equation}
\Delta W_i = -\eta\left(g_i\mathbf{u}_i\mathbf{v}_i^\top+\sigma_i\mathbf{p}_i\mathbf{v}_i^\top+\sigma_i\mathbf{u}_i\mathbf{q}_i^\top\right).
\label{eq:full_update}
\end{equation}
Substituting into~\eqref{eq:benefit},
\begin{align}
\widetilde B_i
&=\mathcal{L}_{\mathrm{new}}(W)-\mathcal{L}_{\mathrm{new}}(W+\Delta W_i)\nonumber\\
&\approx-\left\langle G_{\mathrm{new}},\Delta W_i\right\rangle\nonumber\\
&=\eta\left\langle G_{\mathrm{new}},\,g_i\mathbf{u}_i\mathbf{v}_i^\top+\sigma_i\mathbf{p}_i\mathbf{v}_i^\top+\sigma_i\mathbf{u}_i\mathbf{q}_i^\top\right\rangle\nonumber\\
&=\eta g_i\operatorname{tr}(G_{\mathrm{new}}^\top\mathbf{u}_i\mathbf{v}_i^\top)+\eta\sigma_i\operatorname{tr}(G_{\mathrm{new}}^\top\mathbf{p}_i\mathbf{v}_i^\top)+\eta\sigma_i\operatorname{tr}(G_{\mathrm{new}}^\top\mathbf{u}_i\mathbf{q}_i^\top)\nonumber\\
&=\eta g_i\left(\mathbf{v}_i^\top G_{\mathrm{new}}^\top\mathbf{u}_i\right)+\eta\operatorname{tr}\!\left((\sigma_i G_{\mathrm{new}}\mathbf{v}_i)^\top\mathbf{p}_i\right)+\eta\operatorname{tr}\!\left(\mathbf{q}_i^\top(\sigma_i G_{\mathrm{new}}^\top\mathbf{u}_i)\right)\nonumber\\
&=\eta g_i^2+\eta\operatorname{tr}\!\left((\mathbf{p}_i+\sigma_i(\mathbf{u}_i^\top G_{\mathrm{new}}\mathbf{v}_i)\mathbf{u}_i)^\top\mathbf{p}_i\right)+\eta\operatorname{tr}\!\left(\mathbf{q}_i^\top(\mathbf{q}_i+\sigma_i(\mathbf{v}_i^\top G_{\mathrm{new}}^\top\mathbf{u}_i)\mathbf{v}_i)\right)\nonumber\\
&=\eta\left(g_i^2+\|\mathbf{p}_i\|_2^2+\|\mathbf{q}_i\|_2^2\right).
\label{eq:benefit_final}
\end{align}
Substituting into~\eqref{eq:cost},
\begin{align}
\widetilde C_i
&=\operatorname{tr}\left(\Delta W_i\Sigma_{\mathrm{old}}\Delta W_i^\top\right)\nonumber\\
&=\eta^2\operatorname{tr}\Big[\left(g_i\mathbf{u}_i\mathbf{v}_i^\top+\sigma_i\mathbf{p}_i\mathbf{v}_i^\top+\sigma_i\mathbf{u}_i\mathbf{q}_i^\top\right)\Sigma_{\mathrm{old}}\left(g_i\mathbf{v}_i\mathbf{u}_i^\top+\sigma_i\mathbf{v}_i\mathbf{p}_i^\top+\sigma_i\mathbf{q}_i\mathbf{u}_i^\top\right)\Big]\nonumber\\
&=\eta^2g_i^2\operatorname{tr}\left(\mathbf{u}_i\mathbf{v}_i^\top\Sigma_{\mathrm{old}}\mathbf{v}_i\mathbf{u}_i^\top\right)
+\eta^2g_i\sigma_i\operatorname{tr}\left(\mathbf{u}_i\mathbf{v}_i^\top\Sigma_{\mathrm{old}}\mathbf{v}_i\mathbf{p}_i^\top\right)
+\eta^2g_i\sigma_i\operatorname{tr}\left(\mathbf{u}_i\mathbf{v}_i^\top\Sigma_{\mathrm{old}}\mathbf{q}_i\mathbf{u}_i^\top\right)\nonumber\\
&\quad+\eta^2\sigma_ig_i\operatorname{tr}\left(\mathbf{p}_i\mathbf{v}_i^\top\Sigma_{\mathrm{old}}\mathbf{v}_i\mathbf{u}_i^\top\right)
+\eta^2\sigma_i^2\operatorname{tr}\left(\mathbf{p}_i\mathbf{v}_i^\top\Sigma_{\mathrm{old}}\mathbf{v}_i\mathbf{p}_i^\top\right)
+\eta^2\sigma_i^2\operatorname{tr}\left(\mathbf{p}_i\mathbf{v}_i^\top\Sigma_{\mathrm{old}}\mathbf{q}_i\mathbf{u}_i^\top\right)\nonumber\\
&\quad+\eta^2\sigma_ig_i\operatorname{tr}\left(\mathbf{u}_i\mathbf{q}_i^\top\Sigma_{\mathrm{old}}\mathbf{v}_i\mathbf{u}_i^\top\right)
+\eta^2\sigma_i^2\operatorname{tr}\left(\mathbf{u}_i\mathbf{q}_i^\top\Sigma_{\mathrm{old}}\mathbf{v}_i\mathbf{p}_i^\top\right)
+\eta^2\sigma_i^2\operatorname{tr}\left(\mathbf{u}_i\mathbf{q}_i^\top\Sigma_{\mathrm{old}}\mathbf{q}_i\mathbf{u}_i^\top\right).
\label{eq:cost_9terms}
\end{align}
Reducing each term via \(\mathrm{tr}(\mathbf a\mathbf b^\top M)=\mathbf b^\top M\mathbf a\), and eliminating cross terms with \(\mathbf u_i^\top\mathbf p_i=0\), \(\mathbf v_i^\top\mathbf q_i=0\),
\begin{align}
\operatorname{tr}\left(\mathbf{u}_i\mathbf{v}_i^\top\Sigma_{\mathrm{old}}\mathbf{v}_i\mathbf{u}_i^\top\right) &= (\mathbf{v}_i^\top\Sigma_{\mathrm{old}}\mathbf{v}_i)(\mathbf{u}_i^\top\mathbf{u}_i) = \mathbf{v}_i^\top\Sigma_{\mathrm{old}}\mathbf{v}_i, \nonumber\\
\operatorname{tr}\left(\mathbf{u}_i\mathbf{v}_i^\top\Sigma_{\mathrm{old}}\mathbf{v}_i\mathbf{p}_i^\top\right) &= (\mathbf{v}_i^\top\Sigma_{\mathrm{old}}\mathbf{v}_i)(\mathbf{p}_i^\top\mathbf{u}_i) = 0, \nonumber\\
\operatorname{tr}\left(\mathbf{u}_i\mathbf{v}_i^\top\Sigma_{\mathrm{old}}\mathbf{q}_i\mathbf{u}_i^\top\right) &= (\mathbf{v}_i^\top\Sigma_{\mathrm{old}}\mathbf{q}_i)(\mathbf{u}_i^\top\mathbf{u}_i) = \mathbf{v}_i^\top\Sigma_{\mathrm{old}}\mathbf{q}_i, \nonumber\\
\operatorname{tr}\left(\mathbf{p}_i\mathbf{v}_i^\top\Sigma_{\mathrm{old}}\mathbf{v}_i\mathbf{u}_i^\top\right) &= (\mathbf{v}_i^\top\Sigma_{\mathrm{old}}\mathbf{v}_i)(\mathbf{u}_i^\top\mathbf{p}_i) = 0, \nonumber\\
\operatorname{tr}\left(\mathbf{p}_i\mathbf{v}_i^\top\Sigma_{\mathrm{old}}\mathbf{v}_i\mathbf{p}_i^\top\right) &= (\mathbf{v}_i^\top\Sigma_{\mathrm{old}}\mathbf{v}_i)\|\mathbf{p}_i\|_2^2, \nonumber\\
\operatorname{tr}\left(\mathbf{p}_i\mathbf{v}_i^\top\Sigma_{\mathrm{old}}\mathbf{q}_i\mathbf{u}_i^\top\right) &= (\mathbf{v}_i^\top\Sigma_{\mathrm{old}}\mathbf{q}_i)(\mathbf{u}_i^\top\mathbf{p}_i) = 0, \nonumber\\
\operatorname{tr}\left(\mathbf{u}_i\mathbf{q}_i^\top\Sigma_{\mathrm{old}}\mathbf{v}_i\mathbf{u}_i^\top\right) &= (\mathbf{q}_i^\top\Sigma_{\mathrm{old}}\mathbf{v}_i)(\mathbf{u}_i^\top\mathbf{u}_i) = \mathbf{v}_i^\top\Sigma_{\mathrm{old}}\mathbf{q}_i, \nonumber\\
\operatorname{tr}\left(\mathbf{u}_i\mathbf{q}_i^\top\Sigma_{\mathrm{old}}\mathbf{v}_i\mathbf{p}_i^\top\right) &= (\mathbf{q}_i^\top\Sigma_{\mathrm{old}}\mathbf{v}_i)(\mathbf{p}_i^\top\mathbf{u}_i) = 0, \nonumber\\
\operatorname{tr}\left(\mathbf{u}_i\mathbf{q}_i^\top\Sigma_{\mathrm{old}}\mathbf{q}_i\mathbf{u}_i^\top\right) &= (\mathbf{q}_i^\top\Sigma_{\mathrm{old}}\mathbf{q}_i)(\mathbf{u}_i^\top\mathbf{u}_i) = \mathbf{q}_i^\top\Sigma_{\mathrm{old}}\mathbf{q}_i.
\label{eq:cost_termwise}
\end{align}
Substituting back into~\eqref{eq:cost_9terms} and keeping only the nonzero terms,
\begin{align}
\widetilde C_i
&=\eta^2g_i^2\left(\mathbf{v}_i^\top\Sigma_{\mathrm{old}}\mathbf{v}_i\right)
+2\eta^2g_i\sigma_i\left(\mathbf{v}_i^\top\Sigma_{\mathrm{old}}\mathbf{q}_i\right)
+\eta^2\sigma_i^2\left(\mathbf{v}_i^\top\Sigma_{\mathrm{old}}\mathbf{v}_i\right)\|\mathbf{p}_i\|_2^2
+\eta^2\sigma_i^2\left(\mathbf{q}_i^\top\Sigma_{\mathrm{old}}\mathbf{q}_i\right)\nonumber\\
&=\eta^2[\sigma_i^2\|\mathbf{p}_i\|_2^2\left(\mathbf{v}_i^\top\Sigma_{\mathrm{old}}\mathbf{v}_i\right)+\left(g_i\mathbf{v}_i+\sigma_i\mathbf{q}_i\right)^\top\Sigma_{\mathrm{old}}\left(g_i\mathbf{v}_i+\sigma_i\mathbf{q}_i\right)].
\label{eq:cost_final}
\end{align}

The default scores in~\eqref{eq:benefit_perp_expand}-\eqref{eq:cost_perp_expand} therefore quantify the marginal effect of enabling the singular-vector directions, whereas~\eqref{eq:benefit_final}-\eqref{eq:cost_final} quantify the joint perturbation of the singular value and its associated singular-vector pair for the corresponding ablation.
\section{Datasets}
\label{app:datasets}

\subsection{Continual-Learning Tasks}
\label{app:continual_datasets}

We construct the continual-learning sequences using the eight language tasks from TRACE~\citep{wang2023trace}. TRACE combines challenging and heterogeneous tasks for which pretrained language models generally have substantial room for adaptation. The selected tasks cover specialized domains, multiple languages, code completion, and mathematical reasoning. Following the TRACE protocol, we use 1,000 training examples for each task and retain the corresponding test sets for evaluation. Dataset statistics are summarized in Table~\ref{tab:dataset_statistics}.

\paragraph{Domain-specific tasks.}
This group evaluates adaptation to scientific, financial, and meeting-related domains.
\begin{itemize}
    \item \textbf{ScienceQA (Sin.)}~\citep{lu2022learn} contains questions derived from elementary- and high-school curricula across natural, social, and language science. We retain only the text-based examples, thereby excluding samples that require visual inputs.

    \item \textbf{FOMC (Fom.)}~\citep{shah2023trillion} is a financial stance classification task in which the model determines whether a Federal Reserve policy statement is hawkish or dovish. We use the combined version containing meeting minutes, press-conference transcripts, and speeches.

    \item \textbf{MeetingBank (Mee.)}~\citep{hu2023meetingbank} is a meeting summarization dataset built from city-council proceedings. It requires the model to identify salient information from long transcripts and generate concise summaries.
\end{itemize}

\paragraph{Multilingual tasks.}
The multilingual tasks evaluate whether a model can acquire new information and follow task-specific output formats in languages other than English.
\begin{itemize}
    \item \textbf{C-STANCE (Cst.)}~\citep{zhao2023c} is a Chinese stance detection dataset collected from Sina Weibo. We use its target-based setting, in which the targets appearing in the test set are not observed during training.

    \item \textbf{20Minuten (20M.)}~\citep{gonzales2021new} is a German text simplification dataset derived from the Swiss news outlet \emph{20 Minuten}. Each example pairs a news article with a simplified version, making the task a conditional German text-generation problem.
\end{itemize}

\paragraph{Code-completion task.}
\textbf{Py150}~\citep{lu1codexglue} is constructed from approximately 150,000 Python source files collected from public GitHub repositories. Given a code context, the model must predict the subsequent code content. The task evaluates adaptation under structured inputs and output constraints that differ from ordinary natural-language generation.

\paragraph{Mathematical-reasoning tasks.}
TRACE includes two tasks derived from NumGLUE~\citep{mishra2022numglue}: \textbf{NumGLUE-cm (Ngc.)} and \textbf{NumGLUE-ds (Ngd.)}. Both tasks require numerical computation and multi-step reasoning over natural-language problem statements, providing complementary continual-learning tasks for mathematical reasoning.

\begin{table*}[t]
    \centering
    \caption{Statistics of the continual-learning tasks and capability-preservation benchmarks. A dash indicates that a benchmark is used only for evaluation and is not included in continual training.}
    \label{tab:dataset_statistics}
    \resizebox{\textwidth}{!}{
    \begin{tabular}{lcccccccccc}
        \toprule
        \multirow{2}{*}{Split}
        & \multicolumn{8}{c}{Continual-learning tasks}
        & \multicolumn{2}{c}{Capability-preservation benchmarks} \\
        \cmidrule(lr){2-9}
        \cmidrule(lr){10-11}
        & Sin. & Fom. & Mee. & Cst. & 20M. & Py150
        & Ngc. & Ngd. & HumanEval & GSM8K \\
        \midrule
        Train
        & \multicolumn{8}{c}{1,000 examples per task}
        & -- & -- \\
        Test
        & 2,000 & 496 & 692 & 2,000 & 200 & 2,000
        & 81 & 325 & 164 & 1,319 \\
        \bottomrule
    \end{tabular}
    }
\end{table*}

\subsection{Capability-Preservation Benchmarks}
\label{app:retention_datasets}

We evaluate the preservation of pretrained code-generation and mathematical-reasoning capabilities using HumanEval and GSM8K, respectively. Neither benchmark is included in the continual-training sequences.

\paragraph{HumanEval.}
HumanEval~\citep{chen2021evaluating} contains 164 hand-written Python programming problems. Each problem specifies a function signature and a natural-language docstring, from which the model must generate a complete function implementation. Correctness is determined by executing the generated code against hidden unit tests. We use the full benchmark to measure whether sequential domain adaptation degrades the model's pretrained code-generation capability.

\paragraph{GSM8K.}
GSM8K~\citep{cobbe2021training} is a collection of linguistically diverse grade-school mathematics word problems written by human annotators. Its problems typically require multiple steps of elementary arithmetic together with natural-language reasoning. We evaluate models on the complete official test split of 1,319 problems and report the proportion of problems for which the model produces the correct final answer. GSM8K examples are used only for evaluation, allowing changes in performance to reflect the retention of pretrained mathematical-reasoning capability rather than additional training on the benchmark.

\subsection{General-Domain Reference Corpus}
\label{app:reference_corpus}

SVC estimates the potential forgetting cost of each singular-vector channel using activation statistics collected from an unlabeled general-domain reference corpus. We construct this corpus by sampling 30,000 examples from C4~\citep{raffel2020exploring} and the Python subset of The Stack v2~\citep{lozhkov2024starcoder} at a fixed ratio of 2:1. This corresponds to 20,000 C4 examples and 10,000 Python examples. The sampled corpus is fixed and shared across all models, tasks, and task orders.

The reference corpus is not used to optimize model parameters and does not provide supervision for any continual-learning task. We use only its input texts to collect layer-wise activations and estimate the corresponding second-moment matrices required by the forgetting-cost calculation. Thus, the same reference data can be reused throughout the task sequence without storing examples from previous tasks.
\section{Baselines and Implementation Details}
\label{app:baselines_implementation}
\subsection{Baselines}
\label{app:baselines}

We compare SVC with four groups of continual-learning baselines: regularization and gradient-constraint methods, replay-based methods, parameter-efficient fine-tuning methods, and full or constrained full-parameter fine-tuning methods. Unless otherwise specified, all methods use the same task data, task orders, optimization schedule, and evaluation protocols.

\subsubsection{Regularization and Gradient-Constraint Methods}
\paragraph{Elastic Weight Consolidation.}
Elastic Weight Consolidation (EWC)~\citep{kirkpatrick2017overcoming} penalizes changes to parameters that are estimated to be important for previous tasks. When learning task \(t\), its objective is
\begin{equation}
    \mathcal{L}_{\mathrm{EWC}}
    =
    \mathcal{L}_{t}(\theta)
    +
    \lambda
    \sum_i
    F_i
    \left(\theta_i-\theta^{*}_i\right)^2,
\end{equation}
where \(\theta^{*}\) denotes the parameters obtained after the preceding task and \(F_i\) is the estimated Fisher information for parameter \(i\). We set the regularization coefficient to \(\lambda=0.5\).

\paragraph{Gradient Episodic Memory.}
Gradient Episodic Memory (GEM)~\citep{lopez2017gradient} maintains an episodic memory \(\mathcal{M}_k\) for each previous task \(k\). When learning task \(t\), it computes the current gradient \(g_t\) and a reference gradient \(g_k\) from each \(\mathcal{M}_k\). If \(\langle g_t,g_k\rangle < 0\) for any \(k<t\), GEM replaces \(g_t\) with the solution to
\begin{equation}
\begin{aligned}
    \min_{\widetilde{g}}\quad
    & \frac{1}{2}\|\widetilde{g}-g_t\|_2^2, \\
    \mathrm{s.t.}\quad
    & \langle \widetilde{g},g_k\rangle \geq 0,
    \qquad \forall k<t.
\end{aligned}
\end{equation}
The resulting gradient is used for parameter updates, limiting interference with the stored examples from earlier tasks.

\paragraph{Learning without Forgetting.}
Learning without Forgetting (LwF)~\citep{li2017learning} retains a frozen copy of the preceding model and uses its predictions as soft targets. The training objective combines the current-task loss with a knowledge-distillation term:
\begin{equation}
    \mathcal{L}_{\mathrm{LwF}}
    =
    \mathcal{L}_{t}^{\mathrm{CE}}
    +
    \alpha \mathcal{L}_{\mathrm{KD}}.
\end{equation}
We use a distillation weight of \(\alpha=0.5\) and a softmax temperature of \(T=2\).

\subsubsection{Replay-Based Methods}
Both replay baselines maintain a unified memory buffer containing real examples from previously learned tasks. Samples are drawn uniformly across earlier tasks, and no synthetic data or auxiliary generative model is used.

\paragraph{Replay.}
After training on the current task, Replay performs an additional rehearsal phase using samples drawn from the memory buffer. For each previous task, the buffer stores and replays \(1\%\) of its training examples.

\paragraph{Replay-online.}
Replay-online interleaves historical examples with current-task examples during training. At each stage, approximately \(1\%\) of the stored data are sampled from the same memory buffer and mixed with the current-task mini-batches.

\subsubsection{Parameter-Efficient Fine-Tuning Methods}
\label{app c13}
For SeqLoRA, O-LoRA, PiSSA, and MiLoRA, we use rank \(r=8\), scaling factor \(\alpha=32\), and apply the adapters to the \texttt{q\_proj} and \texttt{v\_proj} modules. SeqLoRA, O-LoRA, and MiLoRA use an adapter dropout of \(0.1\). For PiSSA, we disable adapter dropout following its original implementation. PiSSA and MiLoRA otherwise use the same LoRA configuration and differ primarily in whether their low-rank factors are initialized from the principal or minor singular components.

\paragraph{Sequential LoRA.}
Low-Rank Adaptation (LoRA)~\citep{hu2022lora} parameterizes the update to a frozen weight matrix \(W\) as
\begin{equation}
    W' = W + \frac{\alpha}{r}BA,
\end{equation}
where \(A\) and \(B\) are trainable low-rank factors. Sequential LoRA (SeqLoRA) continually updates the same adapter parameters across the complete task sequence.

\paragraph{Orthogonal LoRA.}
Orthogonal LoRA (O-LoRA)~\citep{wang2023orthogonal} assigns successive tasks to task-specific low-rank subspaces and regularizes these subspaces toward orthogonality. It uses the same LoRA configuration as SeqLoRA, with the orthogonality coefficient set to \(\gamma=0.5\).

\paragraph{PiSSA.}
Principal Singular Values and Singular Vectors Adaptation (PiSSA)~\citep{meng2024pissa} decomposes each pretrained weight matrix through singular value decomposition. The \(r\) largest singular values and their corresponding left and right singular vectors form a rank-\(r\) principal component, which is used to initialize the trainable low-rank factors. The remaining singular components are absorbed into a frozen residual matrix, such that the trainable component and the residual together reconstruct the original weight at initialization. PiSSA uses the same rank, scaling factor, and target modules as the other LoRA-based baselines, but disables adapter dropout.

\paragraph{MiLoRA.}
Minor Singular Component Low-Rank Adaptation (MiLoRA)~\citep{wang2025milora} also decomposes each pretrained weight matrix through singular value decomposition. In contrast to PiSSA, MiLoRA uses the \(r\) smallest singular values and their corresponding left and right singular vectors to initialize the trainable low-rank factors. The remaining principal singular components are kept frozen, allowing adaptation to occur within the less dominant singular subspace. MiLoRA uses the same rank, scaling factor, and target modules as SeqLoRA and PiSSA, and uses the same adapter dropout of \(0.1\) as SeqLoRA.

\subsubsection{Full and Constrained Fine-Tuning Methods}
\paragraph{Sequential Full-Parameter Fine-Tuning.}
Sequential full-parameter fine-tuning (SeqFT) updates all model parameters on each task in sequence without explicit regularization, replay, or subspace constraint.

\paragraph{Orthogonal Subspace Fine-Tuning.}
Orthogonal Subspace Fine-Tuning (OSFT)~\citep{nayak2026sculpting} constrains full-parameter updates through layer-wise singular subspaces. Before learning each task, OSFT decomposes every adapted weight matrix as \(W^{(l)}=U^{(l)}\Sigma^{(l)}V^{(l)\top}\). It then estimates the importance \(I^{(l)}\) of layer \(l\) from the average cosine similarity between that layer's input and output activations on a probing corpus. Layers with higher importance scores are assigned larger protected subspaces.
The fraction of singular directions retained at layer \(l\) is determined by
\begin{equation}
    r^{(l)}
    =
    \mathrm{mrr}
    +
    I^{(l)}
    \left(
        \mathrm{trr}-\mathrm{mrr}
    \right),
\end{equation}
where the normalized importance score \(I^{(l)}\) controls the allocation between the minimum retention ratio \(\mathrm{mrr}\) and the target retention ratio \(\mathrm{trr}\). We follow the original configuration and set \(\mathrm{mrr}=0.1\) and \(\mathrm{trr}=0.8\).

Based on \(r^{(l)}\), OSFT selects the singular vectors associated with the largest singular values to form the protected left and right subspaces, denoted by \(U_{\mathrm{high}}^{(l)}\) and \(V_{\mathrm{high}}^{(l)}\), respectively. Given the current gradient \(G^{(l)}\), OSFT removes the component aligned with both protected subspaces:
\begin{equation}
    G_{\mathrm{proj}}^{(l)}
    =
    G^{(l)}
    -
    U_{\mathrm{high}}^{(l)}
    U_{\mathrm{high}}^{(l)\top}
    G^{(l)}
    V_{\mathrm{high}}^{(l)}
    V_{\mathrm{high}}^{(l)\top}.
\end{equation}
The projected gradient \(G_{\mathrm{proj}}^{(l)}\) is then used to update the full weight matrix.

The original OSFT protocol estimates layer importance using samples from the preceding task. This is incompatible with our setting, in which historical task data cannot be revisited and the original pretraining data are unavailable before the first continual-learning task. We therefore use the same fixed, unlabeled general-domain reference corpus employed by SVC as the probing corpus for OSFT. All other OSFT configurations follow the original method.

\paragraph{Additional implementation details.}
Unless otherwise specified above, all methods follow the common optimization and training settings described in Section~\ref{section es}. SVC is applied to all linear modules in the model. Its trainable singular-vector channels are selected adaptively for each layer and task, without a manually specified global selection ratio.

\subsection{Task Orders}
\label{app:task_orders}
We evaluate every method under three task orders to account for the sensitivity of continual-learning performance to task ordering. Each order contains all eight TRACE tasks, and the same orders are used for every model and baseline. 

\noindent\textbf{Order 1:}
C-STANCE \(\rightarrow\)
FOMC \(\rightarrow\)
MeetingBank \(\rightarrow\)
ScienceQA \(\rightarrow\)
NumGLUE-cm \(\rightarrow\)
NumGLUE-ds \(\rightarrow\)
20Minuten \(\rightarrow\)
Py150.

\noindent\textbf{Order 2:}
NumGLUE-cm \(\rightarrow\)
NumGLUE-ds \(\rightarrow\)
FOMC \(\rightarrow\)
20Minuten \(\rightarrow\)
C-STANCE \(\rightarrow\)
MeetingBank \(\rightarrow\)
ScienceQA \(\rightarrow\)
Py150.

\noindent\textbf{Order 3:}
MeetingBank \(\rightarrow\)
FOMC \(\rightarrow\)
20Minuten \(\rightarrow\)
C-STANCE \(\rightarrow\)
ScienceQA \(\rightarrow\)
NumGLUE-cm \(\rightarrow\)
NumGLUE-ds \(\rightarrow\)
Py150.

\subsection{Motivation Experiment Details}
Figure~\ref{fig:motivation} compares LoRA and SVF under independent adaptation to the eight TRACE tasks. Each run starts from the same Llama3-8B pretrained checkpoint and uses the task data and splits described in Section~\ref{section es} and Appendix~\ref{app:continual_datasets}. LoRA follows the configuration in Appendix~\ref{app c13}, with rank \(r=8\), scaling factor \(\alpha=32\), dropout 0.1, and adapters on q\_proj and v\_proj. SVF updates all singular values in every linear layer while keeping the corresponding left and right singular vectors fixed. Both methods use the training settings in Section~\ref{section es}: Adam, a batch size of 64, five epochs, no warmup, and a learning rate of \(10^{-4}\). Figure~\ref{fig:motivation}(a) reports GSM8K Pass@1 before and after each independent adaptation. Figure~\ref{fig:motivation}(b) reports performance on each task’s test set using the metrics defined in Section~\ref{section es}.

\section{Effect of Expanding PEFT Target Modules}
To examine whether SVC’s advantage can be explained solely by its broader module coverage, we expanded the target modules of PiSSA and MiLoRA from q\_proj and v\_proj to all linear modules. We evaluated both methods on Llama3-8B under Order 1, keeping their other experimental settings unchanged. Table~\ref{tab:peft_all_linear} compares these runs with the corresponding main results. Expanding the target modules reduced HumanEval and GSM8K performance for both methods. PiSSA also showed lower OP and BWT. MiLoRA’s OP increased slightly, from 47.90 to 48.46, while its BWT decreased from \(-1.55\) to \(-1.65\). SVC outperformed both expanded baselines on all four metrics. These results indicate that broader module coverage alone does not account for SVC’s advantage in this setting.

\begin{table}[t]
    \centering
    \caption{Effect of target-module coverage on Llama3-8B under Order 1.
    The default PiSSA and MiLoRA results and the SVC result are reproduced
    from Table 1. Higher is better for all metrics.}
    \label{tab:peft_all_linear}
    \small
    \setlength{\tabcolsep}{5pt}
    \begin{tabular}{llrrrr}
        \toprule
        Method & Target modules & HumanEval & GSM8K & OP & BWT \\
        \midrule
        PiSSA  & \texttt{q\_proj}, \texttt{v\_proj} & 28.05 & 39.42 & 47.81 & $-2.47$ \\
        PiSSA  & All linear                     & 10.37 &  3.56 & 39.80 & $-4.19$ \\
        \midrule
        MiLoRA & \texttt{q\_proj}, \texttt{v\_proj} & 29.27 & 45.49 & 47.90 & $-1.55$ \\
        MiLoRA & All linear                     & 24.39 & 41.55 & 48.46 & $-1.65$ \\
        \midrule
        SVC    & All linear                     & \textbf{30.49} & \textbf{51.55}
               & \textbf{49.62} & $\mathbf{-0.94}$ \\
        \bottomrule
    \end{tabular}
\end{table}

\end{document}